\documentclass{relaxed_system_lab}
\usepackage[T1]{fontenc}
\usepackage{charter}
\usepackage{amsmath,amssymb}
\usepackage{enumitem}
\usepackage{xspace}
\usepackage{listings}
\setlist[itemize]{topsep=3pt,itemsep=2pt,parsep=0pt,leftmargin=1.2em}
\definecolor{metricband}{HTML}{F1EEE8}

\newtcolorbox{rqanswer}[1]{
  enhanced,
  colback=labblue!3!white,
  colframe=labblue!55!black,
  colbacktitle=labblue!10!white,
  coltitle=labblue!55!black,
  fonttitle=\bfseries,
  title={Answer to #1},
  boxrule=.45pt,arc=1mm,
  left=2.5mm,right=2.5mm,top=1.5mm,bottom=1.5mm,
  toptitle=1mm,bottomtitle=1mm,
  before skip=8pt,after skip=8pt
}
\newcommand{\keepsectionstart}{\par
  \ifdim\dimexpr\pagegoal-\pagetotal\relax<8\baselineskip
    \newpage
  \fi}

\title{Think Offline, Act in Real Time:\\Self-Improving Visual Game Agents through Code}
\author[1,2]{Zipeng Qiu}
\author[1]{Shusheng Xu}
\author[1]{Jiale Li}
\author[2]{Yukun Zhou}
\author[2]{Chenyue Li}
\author[1,2]{Wenshuang Li}
\author[1]{Tongkai Yang}
\author[1,2]{Binhang Yuan}
\affiliation[1]{Ant Group}
\affiliation[2]{HKUST}
\abstract{Pre-trained models can read screens and reason about games, yet on an unfamiliar game, an agent still has to learn to play. We ask how an agent can self-improve on such a game from a few of its own attempts, in real-time visual games where the agent sees only the screen and the game keeps running while the agent thinks. We study two kinds of policy that can carry this improvement. In \emph{model as policy}, a model selects actions during play; we adapt real-time chunking so that committed actions keep executing while the model reasons, and train a 2B controller for this interface. In \emph{code as policy}, a model writes, tests, and revises a program between attempts, and the program acts during play. Across 170 GameWorld tasks, with at most eight one-minute executions per run, code as policy reaches 47.80\% mean progress, against 30.90\% for our trained controller, and leads in all five game categories. Improving the controller's weights on a single game, by contrast, yields only modest gains. These results favor letting a model think offline and storing what it learns in a program that acts in real time. Project repository: \url{https://github.com/TabibitoQZP/SIGame}.}

\begin{document}
\maketitle
\begin{figure}[!ht]
  \centering
  \includegraphics[width=0.84\linewidth]{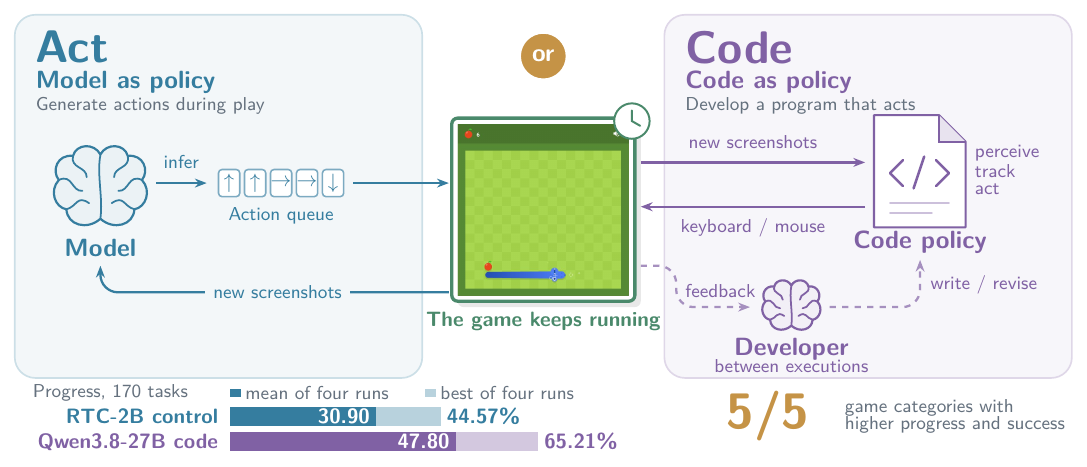}
  \caption{\textbf{Think offline, act in real time.} RTC-2B acts directly in the running game (left), while Qwen3.8-27B develops a program that acts in real time (right). Solid arrows show real-time control; dashed arrows show code development between executions. Bars show progress on 170 tasks: the mean of four runs (dark) and the per-task best of four (light). A direct run contributes one 180-second episode; a code-development run contributes its best submission among up to eight 60-second executions. Code development leads in progress and success in all five categories.}
  \label{fig:teaser}
\end{figure}

\clearpage
\section{Introduction}
\label{sec:introduction}

Pre-trained vision-language models can read a screen and reason about what they see, which makes a single agent that plays many games plausible.
Game foundation models also learn keyboard--mouse or gamepad control directly from large gameplay corpora~\citep{magne2026nitrogen,yue2026openp2p,zhang2026scalinggames}.
On an unfamiliar game, however, such an agent still has to learn how the game responds to its controls, which visual cues matter, and how to repair strategies that fail.
Adapting to a new game usually takes human effort for each game, through game-specific training data, rewards, or hand-written controllers, and that effort does not carry over to the next game.
Self-improvement shifts this work to the agent itself: the agent turns its own attempts into a better policy, as recent agents do when they run their own training experiments~\citep{meng2026rsigym} or refine robot programs from real-world trials~\citep{wang2026pyruareal}.
In \emph{real-time visual games}, where the agent sees only the screen and the game keeps running while it thinks~\citep{zhang2025videogamebench,ouyang2026gameworld}, we ask: \emph{how can an agent self-improve on an unfamiliar game, turning a few of its own attempts into an effective policy?}

An agent that improves itself would reuse one set of capabilities across games, as an experienced player does when picking up an unfamiliar game.
Game-specific systems cannot: traditional real-time game AI relies on rules, controllers, and structured state interfaces built for one game~\citep{ontanon2013survey,churchill2013portfolio}, and OpenAI Five reaches strong Dota~2 play through large-scale self-play with structured observations from a game API~\citep{openai2019dota}.
Moving such a system to another game means redesigning it or training it again.
Real-time games are common, and they make self-improvement harder: every attempt takes real play time, and the game cannot be paused while the agent thinks, so the agent gets few attempts and must keep acting while it decides.

Improvement requires identifying which part of a strategy failed and making a change that works during play.
Consider Snake, shown in Figure~\ref{fig:teaser}.
A collision may result from an incorrect estimate of the head's position, an unsafe turning rule, or a turn that arrives too late.
The same visible failure can therefore call for different revisions to perception, planning, or action timing.
Real-time execution makes these factors interact: a rule that avoids a collision when applied immediately may fail if its action arrives after the snake passes the turning point.
With only a few attempts, the agent has limited opportunity to distinguish these causes and test alternatives.
A revision aimed at the wrong cause wastes an attempt, and repeated local repairs can keep refining an approach that cannot succeed, such as extrapolating pixel motion when the game calls for tracking the grid.
The agent must turn the visual evidence from a failed execution into a policy change that improves the next one.

Two kinds of policy can carry this improvement (Figure~\ref{fig:teaser}).
In \emph{model as policy}, a model selects actions directly during play.
In \emph{code as policy}, a model develops an executable program between attempts, and the program selects actions during play.
The two store what the agent learns in different places: improving a model policy changes its parameters, whereas improving a code policy changes the program that interprets observations and selects actions.
Each leaves a problem open.
A model policy has to reason while the game keeps moving, and improving it on a new game typically means further training, which needs data and rewards for that game.
A code policy can be revised after every attempt, but its program has to recover the game state from pixels on its own.

Prior work handles these problems only in part.
NitroGen and OpenP2P predict game controls from visual observations without generating intermediate reasoning text~\citep{magne2026nitrogen,yue2026openp2p}.
For frontier models that reason before acting, \emph{Scaling in Games} reports that scores on fast-paced 3D tasks drop sharply when the game continues to run during inference~\citep{zhang2026scalinggames}.
Game foundation models also learn by imitating recorded play, without direct feedback on task success or failure.
Code as Policies composes perception and control primitives into robot programs~\citep{liang2023code}, and Voyager, generative code optimization, and GameSense refine game programs from execution feedback, but they observe the game through APIs or object-level features, or keep a VLM in the decision loop~\citep{wang2024voyager,kuang2025gameplaying,lu2025gamesense}.
On GameWorld, ViCoR refines guidance across episodes but still calls a VLM to decide actions during play~\citep{zhou2026vicor}.
The other self-improving agents above work on text and coding benchmarks or robot manipulation.
It remains open which kind of policy makes better use of a few attempts at a running game the agent sees only on screen: a model that keeps deciding during play, or a program that the model revises between attempts and that decides without model calls.

We study both kinds of policy on GameWorld, which contains 170 tasks across 34 games and five categories~\citep{ouyang2026gameworld}.
For model as policy, we adapt real-time chunking (RTC)~\citep{black2025rtc} to discrete game inputs: the model observes recent frames and the actions still queued, and appends a new chunk while the queue executes, so it keeps reasoning without falling behind the game.
We train a 2B controller, RTC-2B, for this interface on games separate from GameWorld, then let GPT-6 Astra~\citep{openai2026astra} lead its further training on a single game, Stack; the gains are modest, although a hand-written program that sees the same screens can complete the simulated tower.
For code as policy, a model writes, tests, and revises a program from screenshots and execution records over a few one-minute executions, and the program estimates the game state from pixels and selects every action during play (Figure~\ref{fig:workflow}).
We further explore how to spend a larger self-improvement budget, comparing independent attempts with developers that review each other's work.
Our results show that, for an agent improving itself on a new real-time game, thinking offline and storing what it learns in a program outperforms direct control in every game category within a few executions, whereas retraining the model on a single game brings only modest gains.

\begin{samepage}
Our contributions are as follows:
\begin{itemize}
  \item \begin{samepage}\emph{A queue-conditioned interface and a reasoning real-time controller.} Timing analysis relates inference delay to action coverage. Trained from Qwen3.5-2B with supervised fine-tuning (SFT) on Qwen3.8-27B's reasoning and actions and then reinforcement learning (RL), RTC-2B reasons before every chunk, matches the mean progress of Qwen3.8-27B on a 31-task subset of GameWorld under the same RTC configuration (21.82\% versus 21.83\%), and leads it in platformer, runner, and simulation games.\par\end{samepage}
  \item \begin{samepage}\emph{Self-improvement through code from a few executions.} With at most eight one-minute executions per run, the best submission of a code-development run averages 47.80\% progress on 170 tasks, against 30.90\% for one RTC-2B episode, with gains in all five game categories. On the 31-task subset, code development also exceeds Qwen3.8-27B playing directly on 20 of the 31 tasks, and a single initial frame suffices to start.\par\end{samepage}
  \item \begin{samepage}\emph{An analysis of how to spend a larger self-improvement budget.} With matched total submission and token caps, the best of four independent runs reaches 65.21\% progress, whereas four developers exchanging peer reviews reach 59.82\%. Traced cases show how peer evidence redirects a stalled developer.\par\end{samepage}
\end{itemize}
\end{samepage}
Section~\ref{sec:model-policy} describes the direct controller, its training, and its further training on Stack.
Section~\ref{sec:code-policy} explains how a developer builds and revises executable policies and how a larger budget can be spent.
Section~\ref{sec:experiments} evaluates both kinds of policy, Section~\ref{sec:related} situates them within prior work, and Section~\ref{sec:conclusion} concludes the paper.

\FloatBarrier
\section{Model as Policy}
\label{sec:model-policy}

This section first defines the interaction setting shared by both kinds of policy, then describes the real-time chunking interface (Section~\ref{sec:rtc}), the training of the controller (Section~\ref{sec:training}), and its further training on a single game (Section~\ref{sec:weight-si}).
Both model and code policies interact with the game through screenshots and keyboard--mouse inputs.
A task specifies the goal, initial configuration, and available controls.
The agent receives public task instructions and may retain its own observations, actions, and timing records; gameplay information, including visible scores, comes from the screenshots.
Internal game state and private evaluator scores are not exposed.
During execution, the game continues to advance through observation, computation, and action dispatch.

A model policy maps the available context to an action or action chunk.
Since each new decision requires inference, the runtime must schedule actions while the next reply is being generated.
We use a queue-conditioned interface for this purpose, train the controller, and then examine its self-improvement on a target game.

\subsection{Real-Time Chunking}
\label{sec:rtc}

RTC overlaps action execution with the generation of the next chunk.
The model plans from both visual observations and the actions already committed to execution, so that its new chunk can follow the remaining queue.

\begin{figure}[t]
  \centering
  \includegraphics[width=\linewidth]{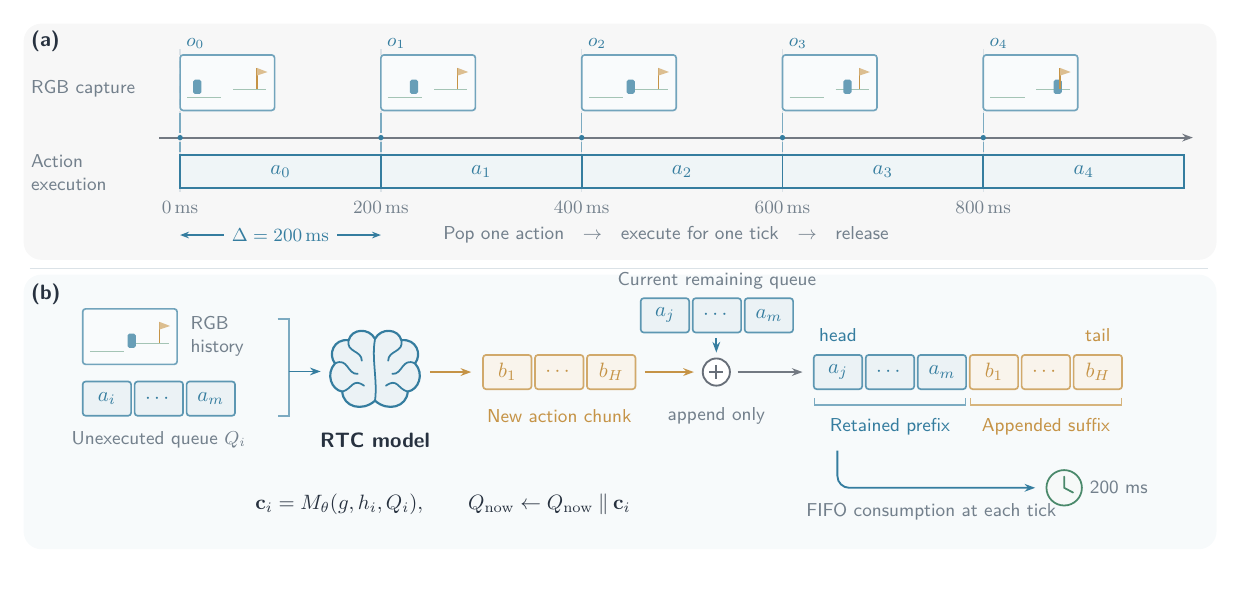}
  \caption{\textbf{RTC with a shared observation and action clock.} (a) Frames are sampled and actions dispatched every 200\,ms. (b) The model receives visual observations and the unexecuted queue, then generates a chunk to append. The queue continues to drain during inference, so the input snapshot differs from the queue remaining at reply time. Blue and gold cells denote retained and newly generated actions.}
  \label{fig:rtc-architecture}
\end{figure}

\paragraph{Observation and execution share a clock.}
We adopt RTC's principle of conditioning on committed actions~\citep{black2025rtc} in a game interface with a fixed tick of $\Delta=200\,\mathrm{ms}$ (Figure~\ref{fig:rtc-architecture}).
The original method uses inpainting in diffusion- or flow-based policies; here the model conditions on a textual queue and generates discrete actions to append.
At each tick, the runtime samples the latest available RGB frame and dispatches the next eligible action from a first-in, first-out queue.
In the standard interface, a keyboard action lasts one tick and is then released; a click is released immediately.
If no queued action is available, the runtime waits while the game continues.
This synchronization concerns observations and action dispatch: model inference proceeds concurrently and need not finish within one tick.

\paragraph{The model plans after the remaining queue.}
For request $i$, let $h_i$ contain the selected visual history, executed-input records, and timing information, and let $Q_i$ be a snapshot of the unexecuted queue.
The model receives the task description $g$ and generates a chunk $\mathbf{c}_i$ of one-tick actions:
\begin{equation}
  \mathbf{c}_i=M_\theta(g,h_i,Q_i),\qquad
  Q_{\mathrm{now}}\leftarrow Q_{\mathrm{now}}\mathbin{\Vert}\mathbf{c}_i.
  \label{eq:rtc-append}
\end{equation}
Here $\Vert$ denotes concatenation, and $Q_{\mathrm{now}}$ is the queue remaining when the valid reply is accepted.
Actions already consumed since the request are not reinserted, and the remaining actions are not replaced or reordered.
A new action becomes eligible only at a tick strictly after receipt.
At most one request is in flight, and invalid replies do not replenish the queue.
These rules make the controls that precede the new chunk explicit to the model.

The visual history describes past interaction, while $Q_i$ specifies future committed inputs.
We combine densely sampled recent frames with sparse older frames so that the model can estimate motion and relate it to its own actions.
Each request reconstructs this context from recorded observations and inputs, rather than extending a conversation containing earlier model reasoning.
Autoregressive robot policies likewise execute actions asynchronously while the next ones are decoded~\citep{lee2026realtimear}.

\paragraph{Latency and action coverage.}
Queue coverage determines both whether execution can continue and when new evidence can affect an action.
Let $L$ be the delay from a request to availability of its valid chunk, $P$ the queued duration at request time, and $C$ the new chunk's duration.
In a periodic regime with fixed latency, fresh observations, one request in flight, a nonempty bounded queue, and no additional scheduling delay, continuous action supply requires $P\ge L$ and $C\ge L$.
The model must therefore plan over a horizon $P+C\ge 2L$, measured from its observation to the end of the new chunk.
Retaining more queued actions covers longer inference, but also postpones the first action that can incorporate new visual evidence.
Appendix~\ref{app:rtc-timing} derives these relations and defines the event-specific response window within which a correction must begin.
Section~\ref{sec:exp-rtc} compares measured latency with the configured response floor $F$.

\subsection{Training the Controller}
\label{sec:training}

We initialize our controller from Qwen3.5-2B.
Qwen3.8-27B supplies the SFT demonstrations and also serves as the direct-control reference and code developer.
We denote the SFT checkpoint as \emph{RTC-2B (SFT)} and its RL successor as \emph{RTC-2B (RL)}; RTC-2B alone refers to the RL checkpoint.

\paragraph{Games and timing.}
We filter 2,522 games from js13kgames~\citep{js13kgames} to obtain 395 self-contained browser games with an identifiable, frequently changing score suitable for training rewards.
These games are separate from the GameWorld evaluation tasks.
A host-side score reader supplies rewards; the policy receives only visual observations, its own action records, the queue, and timing information.
RL rollouts use a generic instruction to discover the controls and make progress, without game-specific task descriptions.
A virtual clock fixes the delay the policy must anticipate during these rollouts: the browser pauses during generation, then advances 3 seconds of committed actions before appending the new chunk.
Evaluation instead keeps the game running throughout inference.
Appendix~\ref{app:training-games} details game selection, and Appendix~\ref{app:training-protocol} specifies the interaction protocol.

\paragraph{Supervised fine-tuning.}
The Qwen3.8-27B teacher plays the training games through the RTC interface.
From its episodes, we retain 31,626 complete requests with syntactically valid action chunks, without filtering by task success.
Each example pairs the visual and queue context with the teacher's full reply, including reasoning and actions.
We update all parameters of Qwen3.5-2B for one epoch of SFT, applying loss only to the target reply.
This trains the model to interpret the queue-conditioned context and produce reasoning followed by an executable chunk.

\paragraph{Reinforcement learning.}
We use a request-level variant of GRPO that assigns a separate advantage to each generated reply.
For each sampled game, we collect eight episodes through the RTC interface.
The host tracks which request produced each executed tick and uses the score change over the interval in which that request's actions execute as its reward.
For each valid request, we sum its reward and those of subsequent requests within the same continuous valid segment.
We compare this return with the mean return of the other episodes at the same request index and normalize within the game group.
The resulting advantage weights the reply's reasoning and action tokens in a clipped policy objective.

Initial experiments with episode-level GRPO showed no clear improvement, so we adopted the request-level objective above.
The resulting RTC-2B (RL) checkpoint is our final controller.
We implement RL training with AReaL~\citep{fu2025areal}, an asynchronous RL framework.
Appendix~\ref{app:rtc-training} gives the reward masks, advantage calculation, and optimization settings.

\subsection{Self-Improvement in Weights}
\label{sec:weight-si}

A model policy improves on a game it has not mastered by changing its weights.
We study this route by further training RTC-2B on a single target game, Stack.
GPT-6 Astra~\citep{openai2026astra} constructs an independent simulator of the game, prepares training data with reasoning and actions generated by Qwen3.8-27B, and trains and evaluates the controller.
Astra revises the procedure from evaluation results, and humans adjust only between stages.
We first apply SFT to specialize RTC-2B to Stack, then train two RL variants from the same SFT checkpoint.
We label the selected SFT checkpoint RTC-2B + Stack SFT and its RL descendants RTC-2B + Stack RL, specifying the reward recipe where needed.
We evaluate the controllers in the simulator and test the SFT and audited-reward RL checkpoints in the continuously running original game.
Section~\ref{sec:exp-training} reports the results; Appendix~\ref{app:training-si} records the simulator, training procedures, and reference controllers.

\FloatBarrier
\section{Code as Policy}
\label{sec:code-policy}

A code policy improves by changing its program rather than a model's weights.
Code development lets the model reason between executions while its program responds to new screenshots during play.
After each execution, the model inspects the recorded outcome and can revise the program.
These two stages have separate budgets: a time limit for each execution, and caps on submissions and model output tokens for development.
Section~\ref{sec:si} follows a policy from the initial evidence through live execution and revision, and Section~\ref{sec:development-search} describes two ways of spending a larger development budget.

\subsection{Self-Improvement through Code}
\label{sec:si}

Code development produces an executable policy that maintains visual state estimates and selects actions throughout an execution.
The developer determines the program's visual processing, stored state, and control logic within a common observation-to-action interface; no task-specific skill decomposition is prescribed.
The generated program can use image-processing libraries and fixed visual services, but the developer model does not plan actions or select skills during execution.

\begin{figure}[t]
  \centering
  \includegraphics[width=\linewidth]{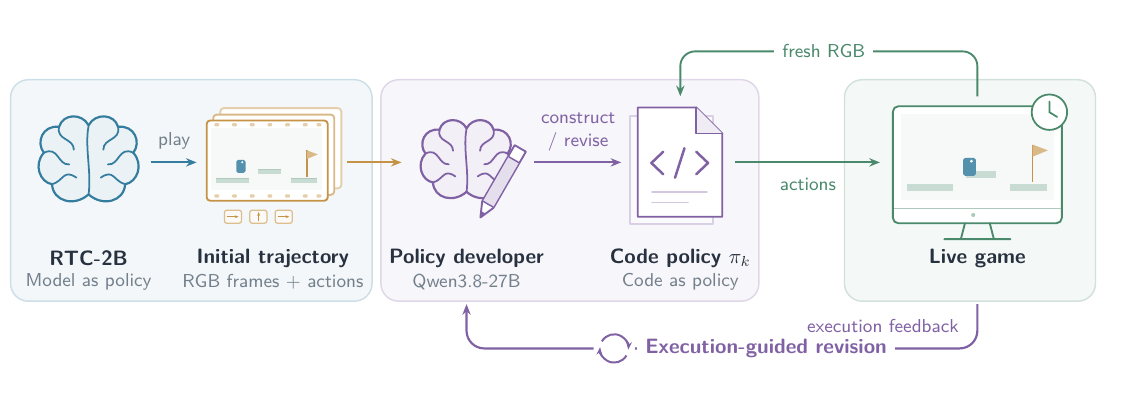}
  \caption{\textbf{Using model interaction to initialize code development.} RTC-2B records visual interaction for a Qwen3.8-27B developer. The developer constructs, tests, and revises a program that processes new screenshots during gameplay. This is the trajectory-initialized workflow evaluated in the main code study; development can also start from a single frame.}
  \label{fig:workflow}
\end{figure}

\paragraph{Initial evidence for code development.}
Code development can begin with either an interaction trajectory or an initial screenshot.
The main evaluation uses a trajectory recorded by RTC-2B, containing screenshots, executed actions, and timestamps (Figure~\ref{fig:workflow}).
Qwen3.8-27B uses these records and the task instructions to construct its first program, then tests and revises it through further executions.
The trajectory provides evidence of how the game responds to inputs; subsequent submissions test the developer's interpretation of that evidence.
Appendix~\ref{app:rtc-timing} details trajectory collection, including its request schedule.
We also evaluate development from the initial screenshot under the same code-development limits to examine the role of initial interaction.

\paragraph{Constructing an executable policy.}
The developer receives the task instructions and the available visual evidence, and can inspect frames, analyze images, and test code offline before submitting a policy.
Offline tests use already collected images; acquiring new gameplay evidence requires a submission.
Submitting freezes the program and its supporting assets for that execution, so later edits cannot change the policy while it plays.
Each submission implements a persistent observation-to-action function,
\begin{equation}
  (a_t,m_{t+1})=\pi_k(o_t,m_t),
  \label{eq:code-policy}
\end{equation}
where $\pi_k$ is the submitted code, $t$ indexes policy calls, $o_t$ is the latest RGB observation at that call, and $m_t$ is the program's own stored state.
The state retains information inferred from earlier frames and is initialized for each submission.
Task knowledge persists across submissions in the code and its supporting assets.
Unlike a precomputed action chunk, the program can choose its next action after processing a new observation.
Image-processing libraries and fixed visual services support this computation, and their runtime counts toward the episode's time limit (Appendix~\ref{app:execution-protocols}); the submitted program combines their outputs into control decisions.

\paragraph{Revising code through execution feedback.}
Each submitted policy executes from the task's prescribed initial configuration.
The developer receives screenshots, executed-input and timing records, and runtime errors, with earlier code versions and execution records available for comparison.
The developer selects which recorded frames and logs to inspect within a continuing development session, and can revise visual processing, action timing, or the decision rule itself.
For example, repeated failure of a reactive rule may motivate maintaining a visual map or planning a sequence of moves.
Feedback includes outcomes visible in recorded frames, but excludes the evaluator's private progress and success signals.
The developer continues testing and revision until it finishes or reaches its execution or output-token allowance.

\paragraph{Executing the program in real time.}
Code policies use the same 200\,ms action clock, with at most one policy call in flight.
Each call returns one tick of action; after a slow call finishes, the next call uses the latest available frame rather than replaying missed observations.
The returned action is queued for a subsequent tick.
New visual evidence reaches control through the program, without a developer-model request for every action.
A program that finishes within one tick can update its decision at the next observation without committing a multi-tick plan to cover model inference.
Code policies therefore use one-action dispatch rather than the RTC chunk-refill procedure.
The program's responsiveness depends on image processing, visual-service calls, and dispatch timing.

\subsection{Allocating the Self-Improvement Budget}
\label{sec:development-search}

Additional submissions and model computation can support more independent solutions or let developers exchange evidence.
Independent histories give each run room to pursue its own approach.
Peer evidence can instead challenge assumptions that persist through local revisions.
We compare four independent runs with a team of four developers that exchange evidence midway, with the same total submission and output-token caps.

\paragraph{Independent exploration.}
Four developers start from the same instructions and initial trajectory, with separate conversations, workspaces, and budgets.
Each tests and revises its code for up to eight submissions without seeing another run's code or execution records.

\paragraph{Independent exploration followed by peer review.}
The collaborative condition introduces one round of peer review after the first four submissions.
Once all branches reach this point or terminate, eligible developers inspect another branch's submitted code and public execution records in a ring and send it a review.
The exchange exposes frozen submitted versions and their visual evidence, while conversations, unsubmitted edits, and private evaluator scores remain separate.
They then continue their own development using the received review and peer snapshots.
Developers can learn both from the implementation they inspect and from the critique they receive.
This schedule lets developers form initial solutions before considering alternatives, while retaining time to test any subsequent changes.
Review computation counts toward the original output-token allowances; total caps match four independent runs (Section~\ref{sec:development-budget}).

\FloatBarrier
\suppressfloats[t]
\keepsectionstart
\section{Experiments}
\label{sec:experiments}

We evaluate both kinds of policy and how each improves, through training for model as policy and through development for code as policy, and then examine how to allocate a larger self-improvement budget.
The experiments address three questions:
\begin{enumerate}[label=\textbf{RQ\arabic*.},leftmargin=3em,itemsep=3pt]
  \item \emph{How well can a model policy act in real time, and how far can it self-improve on a single game?} We examine action supply and latency, broad SFT and RL for a small controller, and its further training on a single game (Sections~\ref{sec:exp-rtc} and~\ref{sec:exp-training}; methods in Section~\ref{sec:model-policy}).
  \item \emph{How effectively can a code policy self-improve from a few executions?} We compare code development with direct control across 170 tasks and with the model held fixed, and examine how development uses initial evidence (Section~\ref{sec:exp-si}; methods in Section~\ref{sec:code-policy}).
  \item \emph{How should a larger self-improvement budget be allocated across independent attempts and peer review?} We compare the two at matched total submission and token caps and trace how peer evidence changes individual developers (Section~\ref{sec:development-budget}; protocol in Section~\ref{sec:development-search}).
\end{enumerate}

\subsection{Benchmark and Metrics}
\label{sec:experimental-setup}

\paragraph{GameWorld.}
GameWorld~\citep{ouyang2026gameworld} contains 34 games across five categories: Arcade, Platformer, Puzzle, Runner, and Simulation.
Each game has five tasks indexed \texttt{01}--\texttt{05}, for 170 tasks in total.
Tasks vary in their target scores, required achievements, or starting levels; \texttt{05} typically specifies one of a game's most demanding objectives.
For detailed comparisons, we use a \emph{31-task subset}: the \texttt{05} task from each of 31 games, spanning all five categories.
This subset supports analyses of execution settings, training, and initialization, as well as a comparison of direct control and code development by the same model.
On the \emph{full 170-task benchmark}, we compare RTC-2B with code policies and study how to allocate additional development attempts.
Appendix~\ref{app:task-coverage} details task selection and scoring.

\paragraph{Models.}
We compare the original Qwen3.5-2B with RTC-2B (SFT) and RTC-2B (RL); RTC-2B without a stage suffix denotes the RL checkpoint.
Qwen3.8-27B serves as the larger direct-control reference and the code developer.
RTC-2B is trained on 395 browser games filtered from js13kgames~\citep{js13kgames}, separate from these evaluation tasks (Appendix~\ref{app:rtc-training}).

\paragraph{Metrics and repeated runs.}
We report task progress (PG) and success rate (SR), both as percentages.
PG is the benchmark's normalized progress toward the task objective; SR is the fraction of runs that achieve it.
For each task, we compute the mean, minimum, and maximum over repeated runs, then average each statistic across tasks.
Max is the mean per-task best of four runs and Min the mean per-task worst.
Tables show the mean above the [Min, Max] pair and break results down by the five GameWorld categories, weighting tasks equally; Overall averages over the tasks covered by each table.
Evaluator scores are used only for reporting and are never shown to policies or developers (Appendix~\ref{app:evaluation}).

\paragraph{Budgets.}
Table~\ref{tab:budgets} lists the allowances of each condition.
Each execution has a wall-clock limit that includes observation, computation, waiting, and action dispatch; restarting the game within an execution does not reset this clock.
The two routes spend interaction differently by design: a model policy plays one continuous 180-second window, while a code-development run spreads up to eight 60-second executions across its revisions, with 256K output tokens for the developer.
By default, code development starts from a trajectory of up to 60 seconds recorded by RTC-2B and shared by all runs of a task.
A four-developer team has the same total caps as four independent runs: 32 submissions and 1M output tokens.

\begin{table}[t]
  \centering
  \small\setlength{\tabcolsep}{5pt}\renewcommand{\arraystretch}{1.12}
  \begin{tabular}{@{}lrrrl@{}}
    \toprule
    Condition & Tasks & Runs & Game time & Development limits \\
    \midrule
    Model policies, subset & 31 & 4 & 180\,s & --- \\
    RTC-2B, full benchmark & 170 & 4 & 180\,s & --- \\
    Independent code, subset & 31 & 4 & $\le$480\,s & 8 submissions, 256K tokens \\
    Independent code, full & 170 & 4 & $\le$480\,s & 8 submissions, 256K tokens \\
    Collaborative code, full & 170 & 1 team & $\le$1,920\,s & 32 submissions, 1M tokens \\
    \bottomrule
  \end{tabular}
  \caption{\textbf{Interaction and development allowances.} Runs are per task and condition; the collaborative condition has one four-branch team per task. Code development uses the same per-run limits on the subset and the full benchmark. Game time is per run (or team), excluding the shared initial trajectory.}
  \label{tab:budgets}
\end{table}

\subsection{Real-Time Execution}
\label{sec:exp-rtc}

A queue keeps actions flowing during inference, but its length also sets how long control runs before a new plan takes effect.
On the 31-task subset, we first compare the GameWorld baseline with two RTC configurations using Qwen3.8-27B, with four 180-second runs per task and condition (Table~\ref{tab:rtc-comparison}).
The baseline uses GameWorld's official generalist adapter, which gives each controlled role up to two previous interaction rounds and the current screenshot, and invokes game-specific semantic actions with their prescribed durations.
RTC instead uses sampled visual history, a 200\,ms tick, and queue-conditioned action chunks.
The 3s and 6s configurations pair response floors of 3 and 6 seconds with 15- and 30-tick chunks; a reply that finishes before the floor $F$ waits until $F$, and a slower reply keeps its actual delay.

\paragraph{The 6s configuration attains the highest mean progress.}
RTC at 6s reaches 26.46\% mean progress and 11.29\% success, against 24.99\% and 7.26\% for the baseline, and its per-task best of four reaches 36.89\%.
Its largest gains come from Platformer (23.03\% versus 10.34\%) and Puzzle (56.67\% versus 46.56\%), while the baseline stays stronger in Arcade and Runner.
The shorter 3s configuration reaches only 21.83\%, below both.
Lowering the floor also shortens the requested chunk, so the outcome depends on whether the queue covers the actual inference delay.

\begin{table}[t]
  \centering
  \small\setlength{\tabcolsep}{4pt}\renewcommand{\arraystretch}{1.08}
  \begin{tabular}{@{}lrrrrr>{\columncolor{labblue!5}}r@{}}
\toprule
Method & Arcade & Platformer & Puzzle & Runner & Simulation & Overall \\
Tasks & 7 & 7 & 6 & 8 & 3 & 31 \\
\midrule
\multicolumn{7}{@{}>{\columncolor{metricband}[0pt][0pt]}l@{}}{\rule[-3pt]{0pt}{15pt}\hspace{4pt}\textbf{Task progress (\%)}} \\
\addlinespace[2pt]
\shortstack[l]{GameWorld\\baseline} & \shortstack[r]{20.21\\[-1pt]{\scriptsize\color{black!65}[16.05, 27.17]}} & \shortstack[r]{10.34\\[-1pt]{\scriptsize\color{black!65}[7.50, 15.00]}} & \shortstack[r]{46.56\\[-1pt]{\scriptsize\color{black!65}[35.76, 55.48]}} & \shortstack[r]{29.66\\[-1pt]{\scriptsize\color{black!65}[19.75, 35.54]}} & \shortstack[r]{14.72\\[-1pt]{\scriptsize\color{black!65}[9.40, 28.38]}} & \shortstack[r]{24.99\\[-1pt]{\scriptsize\color{black!65}[18.25, 32.18]}} \\
\addlinespace[3pt]
RTC ($F=3$s) & \shortstack[r]{13.29\\[-1pt]{\scriptsize\color{black!65}[2.23, 22.43]}} & \shortstack[r]{10.73\\[-1pt]{\scriptsize\color{black!65}[9.71, 13.63]}} & \shortstack[r]{45.52\\[-1pt]{\scriptsize\color{black!65}[31.14, 66.67]}} & \shortstack[r]{20.89\\[-1pt]{\scriptsize\color{black!65}[17.31, 26.60]}} & \shortstack[r]{22.75\\[-1pt]{\scriptsize\color{black!65}[11.11, 47.52]}} & \shortstack[r]{21.83\\[-1pt]{\scriptsize\color{black!65}[14.27, 32.51]}} \\
\addlinespace[3pt]
RTC ($F=6$s) & \shortstack[r]{12.06\\[-1pt]{\scriptsize\color{black!65}[2.23, 26.50]}} & \shortstack[r]{23.03\\[-1pt]{\scriptsize\color{black!65}[21.14, 25.40]}} & \shortstack[r]{56.67\\[-1pt]{\scriptsize\color{black!65}[33.77, 74.59]}} & \shortstack[r]{23.04\\[-1pt]{\scriptsize\color{black!65}[20.67, 26.92]}} & \shortstack[r]{16.79\\[-1pt]{\scriptsize\color{black!65}[5.56, 39.15]}} & \shortstack[r]{26.46\\[-1pt]{\scriptsize\color{black!65}[17.69, 36.89]}} \\
\addlinespace[3pt]
\midrule
\multicolumn{7}{@{}>{\columncolor{metricband}[0pt][0pt]}l@{}}{\rule[-3pt]{0pt}{15pt}\hspace{4pt}\textbf{Success rate (\%)}} \\
\addlinespace[2pt]
\shortstack[l]{GameWorld\\baseline} & \shortstack[r]{0.00\\[-1pt]{\scriptsize\color{black!65}[0.00, 0.00]}} & \shortstack[r]{0.00\\[-1pt]{\scriptsize\color{black!65}[0.00, 0.00]}} & \shortstack[r]{29.17\\[-1pt]{\scriptsize\color{black!65}[16.67, 33.33]}} & \shortstack[r]{6.25\\[-1pt]{\scriptsize\color{black!65}[0.00, 12.50]}} & \shortstack[r]{0.00\\[-1pt]{\scriptsize\color{black!65}[0.00, 0.00]}} & \shortstack[r]{7.26\\[-1pt]{\scriptsize\color{black!65}[3.23, 9.68]}} \\
\addlinespace[3pt]
RTC ($F=3$s) & \shortstack[r]{0.00\\[-1pt]{\scriptsize\color{black!65}[0.00, 0.00]}} & \shortstack[r]{0.00\\[-1pt]{\scriptsize\color{black!65}[0.00, 0.00]}} & \shortstack[r]{20.83\\[-1pt]{\scriptsize\color{black!65}[16.67, 33.33]}} & \shortstack[r]{3.12\\[-1pt]{\scriptsize\color{black!65}[0.00, 12.50]}} & \shortstack[r]{8.33\\[-1pt]{\scriptsize\color{black!65}[0.00, 33.33]}} & \shortstack[r]{5.65\\[-1pt]{\scriptsize\color{black!65}[3.23, 12.90]}} \\
\addlinespace[3pt]
RTC ($F=6$s) & \shortstack[r]{0.00\\[-1pt]{\scriptsize\color{black!65}[0.00, 0.00]}} & \shortstack[r]{14.29\\[-1pt]{\scriptsize\color{black!65}[14.29, 14.29]}} & \shortstack[r]{33.33\\[-1pt]{\scriptsize\color{black!65}[16.67, 50.00]}} & \shortstack[r]{3.12\\[-1pt]{\scriptsize\color{black!65}[0.00, 12.50]}} & \shortstack[r]{8.33\\[-1pt]{\scriptsize\color{black!65}[0.00, 33.33]}} & \shortstack[r]{11.29\\[-1pt]{\scriptsize\color{black!65}[6.45, 19.35]}} \\
\addlinespace[3pt]
\bottomrule
\end{tabular}

  \caption{\textbf{Action generation across game categories.} All conditions use Qwen3.8-27B on the 31-task subset, with four runs per task. Each cell shows Mean above [Min, Max], computed per task and then averaged within the category; the shaded Overall column is task-weighted. The 3s and 6s labels give the configured response floor and matching chunk duration.}
  \label{tab:rtc-comparison}
\end{table}

\paragraph{Action coverage depends on actual latency.}
Measured request times reveal how often inference exceeds the configured floor (Table~\ref{tab:rtc-operating}).
For Qwen3.8-27B, the median request takes 3.91s under the 3s configuration and 4.28s under the 6s configuration, and 71.8\% and 24.8\% of complete replies overrun their floors.
The 3s queue therefore runs dry more often: it is empty on 53.12\% of ticks against 42.62\% at 6s, even though its replies are more often valid (84.54\% versus 73.88\%).
The baseline's median request is shorter, 1.40s, under its different context, action, and output settings.
Action supply depends on chunk duration, request time, and reply validity together, and Appendix~\ref{app:execution-protocols} gives the full configurations.

\begin{table}[t]
  \centering
  \small\setlength{\tabcolsep}{4pt}\renewcommand{\arraystretch}{1.15}
  \begin{tabular}{@{}lrrrrr@{}}
\toprule
\rowcolor{metricband}
Controller & \shortstack{Request (s)\\P50 / P95} & \shortstack{Release (s)\\P50 / P95} & \shortstack{Legal\\(\%)} & \shortstack{Empty\\(\%)} & \shortstack{Output tokens\\P50} \\
\midrule
Qwen3.8-27B baseline & 1.40 / 2.33 & \textemdash & \textemdash & \textemdash & 91 \\
Qwen3.8-27B + RTC (3s) & 3.91 / 10.37 & 3.89 / 10.16 & 84.54 & 53.12 & 272 \\
Qwen3.8-27B + RTC (6s) & 4.28 / 11.99 & 6.00 / 11.52 & 73.88 & 42.62 & 375 \\
Qwen3.5-2B + RTC (3s) & 19.98 / 21.64 & 20.03 / 21.67 & 0.64 & 99.90 & 8,192 \\
RTC-2B (SFT), 3s & 2.83 / 5.14 & 3.00 / 4.99 & 97.86 & 23.56 & 201 \\
RTC-2B (RL), 3s & 2.57 / 5.44 & 3.00 / 5.29 & 97.48 & 24.56 & 224 \\
\bottomrule
\end{tabular}

  \caption{\textbf{Operating conditions on the 31-task subset.} Request and release columns show P50/P95 in seconds; release includes the response-floor wait. Legal is the fraction of complete replies with valid action syntax. Empty is the task-balanced fraction of ticks without a queued action. Output tokens are per request. Measurement details appear in Appendix~\ref{app:latency-measurements}.}
  \label{tab:rtc-operating}
\end{table}

\subsection{Training and Self-Improvement in Weights}
\label{sec:exp-training}

The interface can supply actions only when the model produces valid, useful chunks.
We therefore compare Qwen3.5-2B, RTC-2B (SFT), and RTC-2B (RL) on the 31-task subset, with four 180-second runs per task and condition.
All three use the 3s floor and 15-tick chunks, and the Qwen3.8-27B results above serve as larger-model references.
Appendix~\ref{app:rtc-training} describes the training games, objectives, and schedule.

\paragraph{RTC-2B matches Qwen3.8-27B in mean progress.}
SFT raises mean progress from 5.08\% to 17.68\%, and RL raises it further to 21.82\% (Table~\ref{tab:rtc-training}).
Under the same 3s configuration, Qwen3.8-27B reaches 21.83\%.
RTC-2B also has the higher success rate, 7.26\% versus 5.65\%.

\begin{table}[t]
  \centering
  \small\setlength{\tabcolsep}{4pt}\renewcommand{\arraystretch}{1.08}
  \begin{tabular}{@{}lrrrrr>{\columncolor{labblue!5}}r@{}}
\toprule
Method & Arcade & Platformer & Puzzle & Runner & Simulation & Overall \\
Tasks & 7 & 7 & 6 & 8 & 3 & 31 \\
\midrule
\multicolumn{7}{@{}>{\columncolor{metricband}[0pt][0pt]}l@{}}{\rule[-3pt]{0pt}{15pt}\hspace{4pt}\textbf{Task progress (\%)}} \\
\addlinespace[2pt]
Qwen3.5-2B & \shortstack[r]{2.21\\[-1pt]{\scriptsize\color{black!65}[2.13, 2.23]}} & \shortstack[r]{1.14\\[-1pt]{\scriptsize\color{black!65}[1.14, 1.14]}} & \shortstack[r]{5.48\\[-1pt]{\scriptsize\color{black!65}[1.50, 16.67]}} & \shortstack[r]{12.63\\[-1pt]{\scriptsize\color{black!65}[11.95, 14.13]}} & \shortstack[r]{0.00\\[-1pt]{\scriptsize\color{black!65}[0.00, 0.00]}} & \shortstack[r]{5.08\\[-1pt]{\scriptsize\color{black!65}[4.11, 7.63]}} \\
\addlinespace[3pt]
RTC-2B (SFT) & \shortstack[r]{15.95\\[-1pt]{\scriptsize\color{black!65}[10.31, 21.55]}} & \shortstack[r]{20.69\\[-1pt]{\scriptsize\color{black!65}[9.71, 24.86]}} & \shortstack[r]{14.67\\[-1pt]{\scriptsize\color{black!65}[9.32, 25.13]}} & \shortstack[r]{24.56\\[-1pt]{\scriptsize\color{black!65}[21.78, 28.14]}} & \shortstack[r]{2.34\\[-1pt]{\scriptsize\color{black!65}[0.00, 9.10]}} & \shortstack[r]{17.68\\[-1pt]{\scriptsize\color{black!65}[11.94, 23.49]}} \\
\addlinespace[3pt]
RTC-2B (RL) & \shortstack[r]{13.21\\[-1pt]{\scriptsize\color{black!65}[11.05, 15.53]}} & \shortstack[r]{21.38\\[-1pt]{\scriptsize\color{black!65}[9.71, 27.36]}} & \shortstack[r]{24.13\\[-1pt]{\scriptsize\color{black!65}[13.60, 46.94]}} & \shortstack[r]{26.26\\[-1pt]{\scriptsize\color{black!65}[21.15, 34.34]}} & \shortstack[r]{26.45\\[-1pt]{\scriptsize\color{black!65}[0.00, 50.26]}} & \shortstack[r]{21.82\\[-1pt]{\scriptsize\color{black!65}[12.78, 32.49]}} \\
\addlinespace[3pt]
\shortstack[l]{Qwen3.8-27B\\RTC ($F=3$s)} & \shortstack[r]{13.29\\[-1pt]{\scriptsize\color{black!65}[2.23, 22.43]}} & \shortstack[r]{10.73\\[-1pt]{\scriptsize\color{black!65}[9.71, 13.63]}} & \shortstack[r]{45.52\\[-1pt]{\scriptsize\color{black!65}[31.14, 66.67]}} & \shortstack[r]{20.89\\[-1pt]{\scriptsize\color{black!65}[17.31, 26.60]}} & \shortstack[r]{22.75\\[-1pt]{\scriptsize\color{black!65}[11.11, 47.52]}} & \shortstack[r]{21.83\\[-1pt]{\scriptsize\color{black!65}[14.27, 32.51]}} \\
\addlinespace[3pt]
\shortstack[l]{Qwen3.8-27B\\RTC ($F=6$s)} & \shortstack[r]{12.06\\[-1pt]{\scriptsize\color{black!65}[2.23, 26.50]}} & \shortstack[r]{23.03\\[-1pt]{\scriptsize\color{black!65}[21.14, 25.40]}} & \shortstack[r]{56.67\\[-1pt]{\scriptsize\color{black!65}[33.77, 74.59]}} & \shortstack[r]{23.04\\[-1pt]{\scriptsize\color{black!65}[20.67, 26.92]}} & \shortstack[r]{16.79\\[-1pt]{\scriptsize\color{black!65}[5.56, 39.15]}} & \shortstack[r]{26.46\\[-1pt]{\scriptsize\color{black!65}[17.69, 36.89]}} \\
\addlinespace[3pt]
\midrule
\multicolumn{7}{@{}>{\columncolor{metricband}[0pt][0pt]}l@{}}{\rule[-3pt]{0pt}{15pt}\hspace{4pt}\textbf{Success rate (\%)}} \\
\addlinespace[2pt]
Qwen3.5-2B & \shortstack[r]{0.00\\[-1pt]{\scriptsize\color{black!65}[0.00, 0.00]}} & \shortstack[r]{0.00\\[-1pt]{\scriptsize\color{black!65}[0.00, 0.00]}} & \shortstack[r]{4.17\\[-1pt]{\scriptsize\color{black!65}[0.00, 16.67]}} & \shortstack[r]{0.00\\[-1pt]{\scriptsize\color{black!65}[0.00, 0.00]}} & \shortstack[r]{0.00\\[-1pt]{\scriptsize\color{black!65}[0.00, 0.00]}} & \shortstack[r]{0.81\\[-1pt]{\scriptsize\color{black!65}[0.00, 3.23]}} \\
\addlinespace[3pt]
RTC-2B (SFT) & \shortstack[r]{7.14\\[-1pt]{\scriptsize\color{black!65}[0.00, 14.29]}} & \shortstack[r]{10.71\\[-1pt]{\scriptsize\color{black!65}[0.00, 14.29]}} & \shortstack[r]{0.00\\[-1pt]{\scriptsize\color{black!65}[0.00, 0.00]}} & \shortstack[r]{3.12\\[-1pt]{\scriptsize\color{black!65}[0.00, 12.50]}} & \shortstack[r]{0.00\\[-1pt]{\scriptsize\color{black!65}[0.00, 0.00]}} & \shortstack[r]{4.84\\[-1pt]{\scriptsize\color{black!65}[0.00, 9.68]}} \\
\addlinespace[3pt]
RTC-2B (RL) & \shortstack[r]{0.00\\[-1pt]{\scriptsize\color{black!65}[0.00, 0.00]}} & \shortstack[r]{10.71\\[-1pt]{\scriptsize\color{black!65}[0.00, 14.29]}} & \shortstack[r]{4.17\\[-1pt]{\scriptsize\color{black!65}[0.00, 16.67]}} & \shortstack[r]{9.38\\[-1pt]{\scriptsize\color{black!65}[0.00, 12.50]}} & \shortstack[r]{16.67\\[-1pt]{\scriptsize\color{black!65}[0.00, 33.33]}} & \shortstack[r]{7.26\\[-1pt]{\scriptsize\color{black!65}[0.00, 12.90]}} \\
\addlinespace[3pt]
\shortstack[l]{Qwen3.8-27B\\RTC ($F=3$s)} & \shortstack[r]{0.00\\[-1pt]{\scriptsize\color{black!65}[0.00, 0.00]}} & \shortstack[r]{0.00\\[-1pt]{\scriptsize\color{black!65}[0.00, 0.00]}} & \shortstack[r]{20.83\\[-1pt]{\scriptsize\color{black!65}[16.67, 33.33]}} & \shortstack[r]{3.12\\[-1pt]{\scriptsize\color{black!65}[0.00, 12.50]}} & \shortstack[r]{8.33\\[-1pt]{\scriptsize\color{black!65}[0.00, 33.33]}} & \shortstack[r]{5.65\\[-1pt]{\scriptsize\color{black!65}[3.23, 12.90]}} \\
\addlinespace[3pt]
\shortstack[l]{Qwen3.8-27B\\RTC ($F=6$s)} & \shortstack[r]{0.00\\[-1pt]{\scriptsize\color{black!65}[0.00, 0.00]}} & \shortstack[r]{14.29\\[-1pt]{\scriptsize\color{black!65}[14.29, 14.29]}} & \shortstack[r]{33.33\\[-1pt]{\scriptsize\color{black!65}[16.67, 50.00]}} & \shortstack[r]{3.12\\[-1pt]{\scriptsize\color{black!65}[0.00, 12.50]}} & \shortstack[r]{8.33\\[-1pt]{\scriptsize\color{black!65}[0.00, 33.33]}} & \shortstack[r]{11.29\\[-1pt]{\scriptsize\color{black!65}[6.45, 19.35]}} \\
\addlinespace[3pt]
\bottomrule
\end{tabular}

  \caption{\textbf{RTC training across game categories.} Qwen3.5-2B and both RTC-2B checkpoints share the 3s floor and chunk configuration, with four runs per task on the 31-task subset. The Qwen3.8-27B results are repeated from Table~\ref{tab:rtc-comparison}. Cells show Mean above [Min, Max].}
  \label{tab:rtc-training}
\end{table}

\begin{figure}[t]
  \centering
  \includegraphics[width=\linewidth]{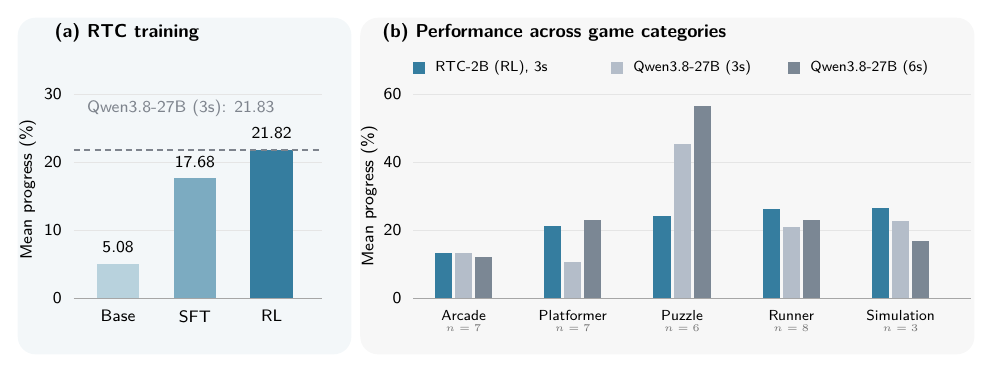}
  \caption{\textbf{RTC training closes the average-progress gap while preserving a distinctive capability profile.} Results use the 31-task subset. (a) Qwen3.5-2B (Base), RTC-2B (SFT), and RTC-2B (RL) under the same 3s configuration; the dashed line marks Qwen3.8-27B with RTC at 3s. (b) Category means reveal strengths in platforming, running, and simulation, together with a remaining Puzzle gap. Categories follow GameWorld, and $n$ is the number of tasks in each.}
  \label{fig:rtc-training}
\end{figure}

\paragraph{Training produces a distinctive control profile.}
SFT improves on the base model in all five categories, and RL adds progress in four, most of all in Simulation (three tasks), from 2.34\% to 26.45\% (Figure~\ref{fig:rtc-training}).
Against Qwen3.8-27B at 3s, RTC-2B is stronger in Platformer (21.38\% versus 10.73\%), Runner (26.26\% versus 20.89\%), and Simulation (26.45\% versus 22.75\%).
Its Runner and Simulation scores also exceed Qwen3.8-27B at 6s, although the larger model keeps the higher overall progress.

Puzzle remains the largest gap: RTC-2B reaches 24.13\% against 45.52\% for Qwen3.8-27B at 3s.
The task-level results associate this gap with search-heavy puzzles, while the trained controller performs well on puzzles requiring fast control (Appendix~\ref{app:task-game-level}).
Similar overall progress thus conceals different strengths in the two models.

\paragraph{SFT teaches the protocol; RL improves the play.}
The operating statistics show how action generation and task performance change with training (Table~\ref{tab:rtc-operating}).
Qwen3.5-2B writes a median of 8,192 tokens per reply and takes 19.98s per request; only 0.64\% of its complete replies are valid, and its queue is empty on 99.90\% of ticks.
After SFT, the median reply shrinks to 201 tokens, validity rises to 97.86\%, and the queue is empty on only 23.56\% of ticks.
RL then raises progress from 17.68\% to 21.82\% while validity (97.48\%) and the empty-queue fraction (24.56\%) stay nearly unchanged.
Its median request is slightly shorter than SFT's (2.57s versus 2.83s) and its P95 slightly longer (5.44s versus 5.14s), and both release at a median of 3.00s under the floor.
The RL gain therefore comes from better action selection within a protocol that SFT has already made reliable.

\paragraph{Improving the weights on Stack yields modest gains.}
Stack requires dropping a moving block onto a tower: any overhang is cut away, reducing the support available for the next layer, and a complete miss ends the game.
Small placement errors therefore accumulate toward the 30-layer goal.
We apply the specialization procedure in Section~\ref{sec:weight-si} and evaluate the resulting weights in the original game (Table~\ref{tab:stack-weight-si}).
Across four games per controller, RTC-2B reaches a mean height of 1.75 layers, compared with 2.25 for specialized SFT and 3.25 for its RL descendant.
RL improves on its SFT parent on three seeds and ties on one, but none of the controllers completes the 30-layer tower.

\paragraph{A visual control program can solve the simulated task.}
In the independent simulator, two RL recipes raise the specialized parent's held-out mean height from 3.50 to 3.75 and 4.00, respectively, after 128 updates each.
A hand-written controller that receives the same RGB requests and knows the simulator's visual conventions completes 14 of 16 games under the same fixed 3-second delay, so an effective program can operate through this interface.
The contrast motivates a further question: can a model develop such visual control programs from interaction, rather than relying on a human to write them?
Appendix~\ref{app:training-si} details the training attempts and reference controllers.

\begin{table}[t]
  \centering
  \small\setlength{\tabcolsep}{6pt}\renewcommand{\arraystretch}{1.12}
  \begin{tabular}{@{}llrrr@{}}
    \toprule
    \rowcolor{metricband}
    Controller & Environment & Games & Mean height & Completed \\
    \midrule
    RTC-2B & Original game & 4 & 1.75 & 0/4 \\
    + Stack SFT & Original game & 4 & 2.25 & 0/4 \\
    + Stack RL, audited reward & Original game & 4 & 3.25 & 0/4 \\
    \midrule
    RTC-2B + Stack SFT & Simulator & 8 & 3.50 & 0/8 \\
    + Stack RL, relative advantage & Simulator & 8 & 3.75 & 0/8 \\
    + Stack RL, audited reward & Simulator & 8 & 4.00 & 0/8 \\
    \midrule
    Hand-written RGB controller & Simulator & 16 & 29.56 & 14/16 \\
    \bottomrule
  \end{tabular}
  \caption{\textbf{Specializing the controller's weights to Stack.} Height is the highest of 30 layers reached in a 180-second game, averaged over games. Original-game rows each use four seeds (42--45); RTC-2B is the Stack reference from the main evaluation. Simulator model rows use eight shared held-out seeds. The hand-written controller is evaluated on 16 seeds with the same RGB interface and fixed 3-second delay.}
  \label{tab:stack-weight-si}
\end{table}

\begin{rqanswer}{RQ1}
\emph{Queue-aware execution and broad training make a small model an effective real-time controller, but improving its weights on a single game yields only modest gains.}
Chunk duration must be assessed against actual latency; SFT makes chunk generation reliable, and RL raises progress within that interface.
RTC-2B reaches 21.82\% mean progress, compared with Qwen3.8-27B's 21.83\% under the same RTC configuration.
On Stack, specialization raises the mean height from 1.75 to 3.25 of 30 layers, while a hand-written RGB controller completes the simulated tower in 14 of 16 games.
\end{rqanswer}

\subsection{Self-Improvement through Code}
\label{sec:exp-si}

We next evaluate whether a model can self-improve through code, developing visual policies through the interface in Section~\ref{sec:code-policy}.
The Qwen3.8-27B developer works in a persistent coding-agent session with file editing, image inspection, and offline Python analysis, and plays the game only through its submissions.
On the full 170-task benchmark, we compare RTC-2B with independent code development initialized from the shared RTC-2B trajectory, with four runs per task for each.
On the 31-task subset, we additionally evaluate development from a single initial frame under the same development limits and compare both initializations with direct control by Qwen3.8-27B.
Each development run reports its best valid submission and whether any submission succeeds.
Captain Callisto's five tasks are assigned zero in every condition (Appendix~\ref{app:evaluation}).

\paragraph{Code development raises progress by 17 points.}
Independent code development reaches 47.80\% mean progress and 27.79\% success across 170 tasks (Table~\ref{tab:code-si}), 16.90 and 15.14 points above direct control by RTC-2B.
A handful of executions, distilled into a program, yields a far stronger policy than direct control.

\begin{table}[t]
  \centering
  \small\setlength{\tabcolsep}{4pt}\renewcommand{\arraystretch}{1.08}
  \begin{tabular}{@{}lrrrrr>{\columncolor{labblue!5}}r@{}}
\toprule
Method & Arcade & Platformer & Puzzle & Runner & Simulation & Overall \\
Tasks & 35 & 40 & 35 & 40 & 20 & 170 \\
\midrule
\multicolumn{7}{@{}>{\columncolor{metricband}[0pt][0pt]}l@{}}{\rule[-3pt]{0pt}{15pt}\hspace{4pt}\textbf{Task progress (\%)}} \\
\addlinespace[2pt]
RTC-2B (RL) & \shortstack[r]{27.40\\[-1pt]{\scriptsize\color{black!65}[21.47, 36.62]}} & \shortstack[r]{22.76\\[-1pt]{\scriptsize\color{black!65}[12.42, 37.41]}} & \shortstack[r]{35.74\\[-1pt]{\scriptsize\color{black!65}[21.75, 53.10]}} & \shortstack[r]{43.41\\[-1pt]{\scriptsize\color{black!65}[37.50, 51.81]}} & \shortstack[r]{19.81\\[-1pt]{\scriptsize\color{black!65}[1.75, 43.35]}} & \shortstack[r]{30.90\\[-1pt]{\scriptsize\color{black!65}[20.85, 44.57]}} \\
\addlinespace[3pt]
\shortstack[l]{Code policy\\Independent} & \shortstack[r]{42.80\\[-1pt]{\scriptsize\color{black!65}[18.83, 63.21]}} & \shortstack[r]{28.76\\[-1pt]{\scriptsize\color{black!65}[13.29, 43.90]}} & \shortstack[r]{70.81\\[-1pt]{\scriptsize\color{black!65}[48.36, 87.83]}} & \shortstack[r]{53.22\\[-1pt]{\scriptsize\color{black!65}[41.64, 66.77]}} & \shortstack[r]{43.51\\[-1pt]{\scriptsize\color{black!65}[16.17, 68.65]}} & \shortstack[r]{47.80\\[-1pt]{\scriptsize\color{black!65}[28.66, 65.21]}} \\
\addlinespace[3pt]
\shortstack[l]{Code policy\\Collaborative} & \shortstack[r]{53.74\\[-1pt]{\scriptsize\color{black!65}[\textemdash,\,\textemdash]}} & \shortstack[r]{36.48\\[-1pt]{\scriptsize\color{black!65}[\textemdash,\,\textemdash]}} & \shortstack[r]{85.07\\[-1pt]{\scriptsize\color{black!65}[\textemdash,\,\textemdash]}} & \shortstack[r]{65.10\\[-1pt]{\scriptsize\color{black!65}[\textemdash,\,\textemdash]}} & \shortstack[r]{62.43\\[-1pt]{\scriptsize\color{black!65}[\textemdash,\,\textemdash]}} & \shortstack[r]{59.82\\[-1pt]{\scriptsize\color{black!65}[\textemdash,\,\textemdash]}} \\
\addlinespace[3pt]
\midrule
\multicolumn{7}{@{}>{\columncolor{metricband}[0pt][0pt]}l@{}}{\rule[-3pt]{0pt}{15pt}\hspace{4pt}\textbf{Success rate (\%)}} \\
\addlinespace[2pt]
RTC-2B (RL) & \shortstack[r]{8.57\\[-1pt]{\scriptsize\color{black!65}[5.71, 17.14]}} & \shortstack[r]{15.00\\[-1pt]{\scriptsize\color{black!65}[7.50, 25.00]}} & \shortstack[r]{12.86\\[-1pt]{\scriptsize\color{black!65}[2.86, 28.57]}} & \shortstack[r]{15.62\\[-1pt]{\scriptsize\color{black!65}[10.00, 25.00]}} & \shortstack[r]{8.75\\[-1pt]{\scriptsize\color{black!65}[0.00, 20.00]}} & \shortstack[r]{12.65\\[-1pt]{\scriptsize\color{black!65}[5.88, 23.53]}} \\
\addlinespace[3pt]
\shortstack[l]{Code policy\\Independent} & \shortstack[r]{21.43\\[-1pt]{\scriptsize\color{black!65}[8.57, 40.00]}} & \shortstack[r]{18.12\\[-1pt]{\scriptsize\color{black!65}[5.00, 30.00]}} & \shortstack[r]{51.43\\[-1pt]{\scriptsize\color{black!65}[28.57, 68.57]}} & \shortstack[r]{23.75\\[-1pt]{\scriptsize\color{black!65}[15.00, 37.50]}} & \shortstack[r]{25.00\\[-1pt]{\scriptsize\color{black!65}[0.00, 55.00]}} & \shortstack[r]{27.79\\[-1pt]{\scriptsize\color{black!65}[12.35, 44.71]}} \\
\addlinespace[3pt]
\shortstack[l]{Code policy\\Collaborative} & \shortstack[r]{25.71\\[-1pt]{\scriptsize\color{black!65}[\textemdash,\,\textemdash]}} & \shortstack[r]{25.00\\[-1pt]{\scriptsize\color{black!65}[\textemdash,\,\textemdash]}} & \shortstack[r]{65.71\\[-1pt]{\scriptsize\color{black!65}[\textemdash,\,\textemdash]}} & \shortstack[r]{42.50\\[-1pt]{\scriptsize\color{black!65}[\textemdash,\,\textemdash]}} & \shortstack[r]{40.00\\[-1pt]{\scriptsize\color{black!65}[\textemdash,\,\textemdash]}} & \shortstack[r]{39.41\\[-1pt]{\scriptsize\color{black!65}[\textemdash,\,\textemdash]}} \\
\addlinespace[3pt]
\bottomrule
\end{tabular}

  \caption{\textbf{Code development on the full 170-task benchmark.} RTC-2B is the RL checkpoint; Qwen3.8-27B develops the code policies. RTC-2B and independent development each have four runs per task, shown as Mean above [Min, Max]. Collaboration reports the best submission of one four-developer team per task, whose total caps match independent Max.}
  \label{tab:code-si}
\end{table}

\paragraph{The gain extends beyond puzzles.}
Code development leads in progress and success in all five categories.
Puzzle shows the largest progress gain, from 35.74\% to 70.81\%, followed by Simulation.
Arcade, Platformer, and Runner, which require continuous reactive control, improve as well.
Appendix~\ref{app:task-game-level} resolves the comparison by task and by game.

\paragraph{Execution feedback improves the program within a run.}
An Edge Surf run shows how the gain arises.
With a target of 800 points, the initial RTC-2B trajectory reaches 485; the developer's first program reaches 751, and its revision reaches 893, completing the task before a later collision.
The program processes the scene to choose keyboard actions, and each execution provides visual evidence for the developer's next revision.
Appendix~\ref{app:2048-code-policy} examines a generated 2048 policy at the level of its code: the first submitted program reads the board from image patches and chooses moves with a median round trip of 5.47\,ms, well within one 200\,ms action tick.

\paragraph{The lead persists with the model fixed.}
The 170-task comparison pairs a Qwen3.8-27B developer with RTC-2B.
On the 31-task subset, we also compare code development with the same Qwen3.8-27B model playing directly (Table~\ref{tab:same-model}).
Against its strongest direct configuration, RTC at 6s, mean progress is 32.30\% for trajectory-initialized development and 32.26\% for a single-frame start, compared with 26.46\% for direct control.
Trajectory-initialized development has higher mean progress on 20 of the 31 tasks and lower on 8; the paired mean difference of $+5.84$ points has a descriptive 95\% interval of $[-1.60, 13.15]$.
The per-task best of four rises from 36.89\% to 50.34\% and 48.00\%, a paired difference whose interval excludes zero (Appendix~\ref{app:same-model}).

\begin{table}[t]
  \centering
  \small\setlength{\tabcolsep}{8pt}\renewcommand{\arraystretch}{1.12}
  \begin{tabular}{@{}lrr@{}}
\toprule
\rowcolor{metricband}
Policy & PG (\%) & SR (\%) \\
\midrule
RTC-2B (RL), direct (3s) & \shortstack[r]{21.82\\[-1pt]{\scriptsize\color{black!65}[12.78, 32.49]}} & \shortstack[r]{7.26\\[-1pt]{\scriptsize\color{black!65}[0.00, 12.90]}} \\
Qwen3.8-27B, GameWorld baseline & \shortstack[r]{24.99\\[-1pt]{\scriptsize\color{black!65}[18.25, 32.18]}} & \shortstack[r]{7.26\\[-1pt]{\scriptsize\color{black!65}[3.23, 9.68]}} \\
Qwen3.8-27B + RTC (3s) & \shortstack[r]{21.83\\[-1pt]{\scriptsize\color{black!65}[14.27, 32.51]}} & \shortstack[r]{5.65\\[-1pt]{\scriptsize\color{black!65}[3.23, 12.90]}} \\
Qwen3.8-27B + RTC (6s) & \shortstack[r]{26.46\\[-1pt]{\scriptsize\color{black!65}[17.69, 36.89]}} & \shortstack[r]{11.29\\[-1pt]{\scriptsize\color{black!65}[6.45, 19.35]}} \\
\midrule
Qwen3.8-27B code, initial frame & \shortstack[r]{32.26\\[-1pt]{\scriptsize\color{black!65}[18.06, 48.00]}} & \shortstack[r]{11.29\\[-1pt]{\scriptsize\color{black!65}[3.23, 19.35]}} \\
Qwen3.8-27B code, RTC trajectory & \shortstack[r]{32.30\\[-1pt]{\scriptsize\color{black!65}[12.87, 50.34]}} & \shortstack[r]{14.52\\[-1pt]{\scriptsize\color{black!65}[3.23, 25.81]}} \\
\bottomrule
\end{tabular}

  \caption{\textbf{One Qwen3.8-27B model used two ways.} Direct-control and code-development rows each have four runs per task on the 31-task subset, shown as Mean above [Min, Max]; RTC-2B (RL) is included for reference. A direct run plays one 180-second episode, whereas a code-development run has up to eight 60-second executions.}
  \label{tab:same-model}
\end{table}

\paragraph{One frame is enough to start.}
To examine the role of initial interaction, we give the developer only the first screenshot on the 31-task subset, retaining the same instructions, tools, and development limits.
We compare four runs per task with the trajectory-initialized runs from the main study (Table~\ref{tab:code-initialization}).
Mean progress is nearly identical: 32.26\% from the frame and 32.30\% from the trajectory.
Trajectory-initialized runs have slightly higher success (14.52\% versus 11.29\%) and per-task best progress (50.34\% versus 48.00\%), and use slightly fewer submissions, 5.50 versus 5.98 among the 123 pairs with known counts.
Development can therefore start without a controller rollout and acquire useful evidence through its own submissions (Appendix~\ref{app:initialization}).

\begin{table}[t]
  \centering
  \small\setlength{\tabcolsep}{12pt}\renewcommand{\arraystretch}{1.15}
  \begin{tabular}{@{}lrrr@{}}
\toprule
\rowcolor{metricband}
Initial evidence & PG (\%) & SR (\%) & \shortstack{Submissions\\Mean} \\
\midrule
Initial frame & \shortstack[r]{32.26\\[-1pt]{\scriptsize\color{black!65}[18.06, 48.00]}} & \shortstack[r]{11.29\\[-1pt]{\scriptsize\color{black!65}[3.23, 19.35]}} & 5.98 \\
Initial RTC trajectory & \shortstack[r]{32.30\\[-1pt]{\scriptsize\color{black!65}[12.87, 50.34]}} & \shortstack[r]{14.52\\[-1pt]{\scriptsize\color{black!65}[3.23, 25.81]}} & 5.50 \\
\bottomrule
\end{tabular}

  \caption{\textbf{Initial evidence for code development.} Both conditions use the 31-task subset, with four runs per task. PG/SR show task-averaged Mean [Min, Max]. Submission means use the 123 task--run pairs with counts available on both sides and exclude the initial rollout.}
  \label{tab:code-initialization}
\end{table}

\begin{rqanswer}{RQ2}
\emph{Within a few executions, self-improvement through code yields policies that outperform direct control in all five game categories.}
On 170 tasks, code development reaches 47.80\% progress and 27.79\% success, against 30.90\% and 12.65\% for RTC-2B.
On the 31-task subset, it also outperforms the strongest direct configuration of the same Qwen3.8-27B model on 20 of the 31 tasks, and a single-frame start reaches nearly the same mean as trajectory initialization.
\end{rqanswer}

\subsection{Allocating the Self-Improvement Budget}
\label{sec:development-budget}

Repeated runs improve the chance of finding a good policy, which raises the question of how a larger self-improvement budget is best spent: on more independent attempts, or on developers that learn from each other, as work on multi-agent debate and collaboration suggests (Section~\ref{sec:development-search}).
On the full 170-task benchmark, we compare the four independent runs with one four-developer collaborative team per task, in which developers exchange code and execution evidence midway through development.
Both conditions allow at most 32 submissions and 1M output tokens in total, and both report the best progress and any success across the valid submissions of their four developers.

\paragraph{Additional independent runs extend task coverage.}
The best of four independent runs reaches 65.21\% mean progress, and at least one run succeeds on 44.71\% of tasks, against 47.80\% and 27.79\% for a single run.
For reference, GameWorld reports 64.1\% progress and 55.3\% success for a novice human player new to the games, under its own protocol of 100 primitive actions per task~\citep{ouyang2026gameworld}.
Averaging over every subset of the four runs for each task traces the gain from one to four runs (Table~\ref{tab:development-budget}): the second run adds 10.04 points of best-attained progress, and the third and fourth add 4.59 and 2.78, for 17.41 points in total.

\begin{table}[t]
  \centering
  \small\setlength{\tabcolsep}{10pt}\renewcommand{\arraystretch}{1.15}
  \begin{tabular}{lrrrr}
\toprule
Independent runs & 1 & 2 & 3 & 4 \\
\midrule
\rowcolor{metricband}
Best-of-$k$ progress (\%) & 47.80 & 57.84 & 62.44 & 65.21 \\
Any-success rate (\%) & 27.79 & 36.76 & 41.62 & 44.71 \\
\bottomrule
\end{tabular}

  \caption{\textbf{Performance attained with additional independent runs.} For each of 170 tasks, we average the maximum PG and any-success indicator over all size-$k$ subsets of its four runs, then average across tasks. Each run is a complete development attempt with up to eight submissions.}
  \label{tab:development-budget}
\end{table}

\paragraph{Independent runs outperform a peer-reviewing team.}
Under the same total submission and token caps, a four-developer team reaches 59.82\% progress and 39.41\% success, well above a single run but below the independent best of four, 65.21\% and 44.71\%.
Independent runs lead in progress in all five categories; only in Runner success does the team come out ahead, 42.50\% versus 37.50\%.

\begin{figure}[t]
  \centering
  \includegraphics[width=\linewidth]{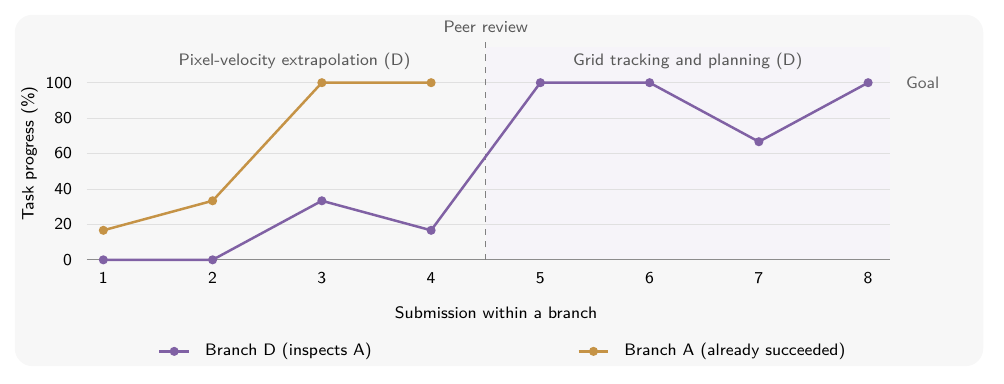}
  \caption{\textbf{Peer inspection redirects a Snake policy.} D reviews A after submission 4, replaces its tracking and planning approach, and reaches the six-apple goal on submission 5; A has already succeeded on submission 3. Curves show the progress of each submission, including regressions. Progress is the apple count divided by the six-apple goal, capped at 100\%. The dashed line marks peer review.}
  \label{fig:snake-development}
\end{figure}

\paragraph{Peer evidence can still redirect a stalled developer.}
Review records and submitted code show how a peer's work can change a developer's state representation and control strategy, as in a Snake task requiring six apples (Figure~\ref{fig:snake-development}).
Developer D starts with pixel-based head detection, velocity estimates, and short-horizon extrapolation.
After changes to direction voting and safety margins, its first four submissions reach only 0, 0, 2, and 1 apples.
When reviewing developer A, D inspects A's implementation and recorded frames, identifies grid tracking and planning as the better approach, and adopts it.
Its next program tracks the head from changes in a visually estimated occupancy grid, plans a path to the apple with breadth-first search, and falls back on flood fill to stay alive.
Its fifth submission reaches six apples, and the remaining three reach six, four, and six.

D learned by inspecting another implementation; it received no valid review of its own.
Because A had already succeeded, the exchange spread a working approach to a stalled developer.
In a complementary NS-Shaft case, an incoming critique instead corrects a rule that waits on a supporting platform, and further debugging then completes the task (Appendix~\ref{app:development-cases}).
The two cases show complementary ways to revise a persistent assumption: demonstrating a working alternative and diagnosing an existing rule.
In aggregate, four independent runs attain higher best scores than the peer-reviewing team, while these cases point to uses of peer evidence that a better exchange schedule could exploit.

\begin{rqanswer}{RQ3}
\emph{Independent attempts make better use of a larger self-improvement budget than a midway round of peer review.}
With the same total submission and token caps, the best of four independent runs reaches 65.21\% progress against 59.82\% for a team that exchanges peer reviews midway.
Peer evidence can redirect a stalled developer, but independent histories reach the higher best result.
\end{rqanswer}

\section{Related Work}
\label{sec:related}

We review real-time visual control, then executable policies and self-improvement, and finally how multiple developers share experience.

\paragraph{Visual game agents and real-time execution.}
NitroGen and OpenP2P train visual controllers from large gameplay corpora~\citep{magne2026nitrogen,yue2026openp2p}.
VideoGameBench compares real-time play with a paused variant to examine inference latency~\citep{zhang2025videogamebench}, while GameWorld provides standardized tasks and verifiable metrics across diverse games, including continuously running environments~\citep{ouyang2026gameworld}.
Scaling in Games studies the effects of training data and model scale, and shows that real-time performance depends on both gameplay capability and inference latency~\citep{zhang2026scalinggames}.
In robotics, real-time chunking overlaps action execution with generation while preserving committed actions~\citep{black2025rtc}; FASTER analyzes how generation and execution horizons affect response delay~\citep{lu2026faster}.
We adapt the committed-action principle to discrete game inputs and train a controller to plan from visual history and the remaining queue, then measure both its latency and its task performance.

\paragraph{Executable policies and self-improvement.}
Self-improving agents run their own improvement experiments on training data and harnesses~\citep{meng2026rsigym}, and coding agents refine programs from execution feedback.
Experience from earlier attempts can also change a model's instructions instead of its weights or code: in TORSI, a slow teacher writes guidance that a fast student follows during live play~\citep{pan2026world}, and ViCoR refines guidance across GameWorld episodes while a VLM still decides actions during play~\citep{zhou2026vicor}.
Programs allow a model's reasoning to guide subsequent actions without regenerating each decision.
Code as Policies composes perception outputs and control primitives into programs~\citep{liang2023code}, and Voyager refines reusable Minecraft skills using game-API observations and execution feedback~\citep{wang2024voyager}.
Cradle and GameSense develop visual control code within modular agent systems~\citep{tan2024cradle,lu2025gamesense}; GameSense's execution modules can contain feedback loops and sustain gameplay under high-level VLM planning.
In Gym-Anything-RT, frontier agents allowed to program write code that takes over real-time perception and action~\citep{aggarwal2026gymanythingrt}.
Generative code optimization revises game policies from traces and language feedback using object-level OCAtari observations~\citep{kuang2025gameplaying}, while Real-Time Reasoning Gym also studies generated code policies operating on structured state~\citep{wen2025realtime}.
In robotics, PyRUA-REAL develops manipulation programs from interaction and freezes them so that most execute without runtime model calls~\citep{wang2026pyruareal}.
Build, Don't Play similarly separates building an agent from playing directly, using a simulator and PPO training to construct a Hearthstone Battlegrounds AI~\citep{ma2026build}.
We study self-improvement of a visual policy from a few attempts at a continuously running game.
The developer constructs the program's visual state estimates and control rules from screenshots and execution records, without internal-state access, and revises them between executions; the program selects every action during play, and we compare it with direct control.

\paragraph{Independent development and collaboration.}
Multi-agent debate uses proposals and critiques to improve answers~\citep{du2024debate}, while ChatDev and MetaGPT organize software development through roles and communication~\citep{qian2024chatdev,hong2024metagpt}.
However, re-evaluations find that debate does not consistently outperform self-consistency or ensembling~\citep{smit2024mad}.
Repeated sampling offers another way to increase solution coverage, with candidate selection determining how much of that benefit can be realized~\citep{brown2024monkeys}.
We compare communication with independent exploration for executable game policies, in which developers exchange code and execution evidence under equal total submission and token caps.

\section{Conclusion}
\label{sec:conclusion}

We study how an agent can self-improve on an unfamiliar game, turning a few of its own real-time visual attempts into an effective policy.
For model as policy, RTC makes committed actions explicit during inference, and broad training yields a small controller that matches the larger reference model's mean progress on the 31-task subset under the same RTC configuration; improving its weights on a single game, however, yields only modest gains.
Code as policy improves from a few executions instead: execution feedback guides revisions to a program that acts on subsequent frames.
On GameWorld, this approach attains higher progress across all five game categories, and its mean advantage remains when the model is held fixed.

With matched total submission and token caps, independent attempts reach higher best scores than developers that review each other's work midway, although peer evidence can redirect a stalled developer.
Together, these findings show how reasoning and interaction feedback can be stored in an executable policy.
The model investigates failures and revises its strategy between attempts, while the program performs visual estimation and control during play.
In this design, real-time constraints apply to the program's action execution, while the reasoning that improves the policy takes place between executions.

\clearpage
\setlength{\bibsep}{2pt}
\bibliographystyle{unsrtnat}
\bibliography{references}
\clearpage
\appendix
\section{Additional Evaluation Details}
\label{app:evaluation}

This appendix specifies task coverage, execution protocols, development resources, score aggregation, and the measurements supporting the main comparisons, together with scoring corrections and initialization details.

\subsection{Task Coverage and Scoring}
\label{app:task-coverage}
The 31-task subset used for execution, training, same-model, and initialization comparisons contains the designated \texttt{\_05} task from each of 31 games across all five categories.
The input interface used for these comparisons did not support sustained mouse-button presses for mining in Minecraft or right-clicks for flagging in Minesweeper, so both games were excluded from the subset.
Captain Callisto was excluded because of runtime/scoring problems.
Task \texttt{\_05} typically specifies a demanding objective within a game, but its index does not define a difficulty level shared across games.
The full 170-task benchmark compares RTC-2B with code policies and independent with collaborative development; it includes Minecraft and Minesweeper with support for these inputs.
Captain Callisto's five tasks are assigned zero in all conditions while remaining in the 170-task denominator.
Execution failures likewise remain in the fixed denominator with zero aggregate score.

The Qwen3.5-2B RTC condition contains 124 runs and is also used in the supplementary baseline comparison.
The full-benchmark RTC-2B reference includes the 124 runs from the 31-task subset; the trajectory-initialized code comparisons likewise reuse the corresponding development runs from the full benchmark.
These subset results are reused observations, not additional independent evaluations.
All comparisons use unrounded task progress and the recorded success indicator.

Two collaborative Fireboy and Watergirl tasks received false success scores during loading.
The main table retains the benchmark-reported scores with the Captain Callisto convention above.
Conservatively setting both affected task scores to zero, while retaining all 170 tasks, changes collaborative PG/SR from 59.82\%/39.41\% to 58.65\%/38.24\%.
The ordering relative to independent development's mean and best-of-four results is unchanged.

\subsection{Execution and Development Protocols}
\label{app:execution-protocols}

\paragraph{Official model-policy baseline.}
The baseline retains GameWorld's coordinator, role-specific controls, and semantic-action execution.
Each role receives up to two previous interaction rounds and its current screenshot; earlier assistant responses remain in the history, but their separate reasoning traces are not reinserted.
The model emits XML tool calls to the game's registered actions, whose durations follow the official action definitions.
Inference does not pause the game, and the 180-second limit begins at the first frame, includes resets, and permits earlier termination on success or a terminal state.
Generation uses temperature 0 and a maximum of 2,048 output tokens per request, with thinking enabled and medium reasoning effort.
There is no imposed response floor or common 200\,ms action grid for this baseline.

\paragraph{RTC model-policy evaluation.}
RTC uses temperature 0.6, top-$p$ 0.95, and a maximum of 8,192 output tokens per request, with thinking enabled and medium reasoning effort.
The 3s and 6s configurations request exactly 15 and 30 action lines and trigger refill when at most 16 and 31 actions remain, respectively.
The model can combine up to four keys and a normalized-coordinate click within one tick; a reset may occur only on the final line of a chunk.
The 180-second limit includes inference and response-floor waits.
Context construction and the separate initial-trajectory schedule are specified in Appendix~\ref{app:rtc-timing}.
The baseline samples evaluator state at official execution steps, whereas RTC records it on its tick schedule.
Both use host-side scores for reporting without exposing them to the policy; the differing sampling schedules are another part of the system-level comparison.

\paragraph{Development environment and evidence.}
We use the Claude Code client with Qwen3.8-27B as its developer model.
The client maintains a persistent conversation with its native context compaction, together with a workspace for code, assets, and execution records.
The developer can read and edit files, inspect images, and run local Python analyses on existing evidence.
Only the submission tool starts a new game execution; a separate status tool reports the remaining allowances.
Each submission freezes the current policy and assets and returns paths to recorded frames, input events, timing, and errors, together with a small set of frame previews.
The developer chooses what to inspect; the entire recorded trajectory is not automatically inserted into its next model request.

The submitted Python function receives an RGB array and its previous local state and returns one action and updated state.
Its SDK additionally exposes its own observation timing, input receipts, queue, and reset identifier, but no internal game variables.
Local state starts as \texttt{None} for each submission and persists across in-run resets; the program can detect a changed reset identifier and clear its state explicitly.
NumPy and OpenCV support local image processing.
Optional visual services provide OCR, segmentation, depth, and fixed-choice judgments over supplied images, without an unrestricted generative-model interface for choosing actions.
Their elapsed time counts toward the running episode.
An in-run reset invalidates pending actions and results from the previous game instance without extending the deadline.
At the deadline, the runtime releases active inputs and discards unexecuted actions.

\paragraph{Budget accounting.}
The output-token caps are 262,144 per developer and 1,048,576 across four developers, abbreviated as 256K and 1M in the main text.
The 256K-token allowance counts successful responses with confirmed usage, including thinking, auxiliary client calls, context compaction, and peer review.
The client permits at most 32,768 output tokens per request, reduced when less budget remains.
Failed requests and requests with unknown usage are recorded separately and do not consume this allowance, which therefore counts accounted output rather than total backend computation or retries.
Every submission attempt consumes one of the eight opportunities, including failed executions.
Peer review permits up to 8,192 output tokens within the branch's original allowance and does not grant additional gameplay.
During review, developers can inspect the frozen peer evidence and write a report, but cannot edit policies or submit executions.

\subsection{Interpreting the Development Comparison}
Comparisons between model policies and code development compare complete systems, each under its own interface and budget.
A direct-control run is one 180-second episode, whereas an independent development run has up to eight 60-second executions and 256K output tokens, and the systems also differ in context, action interface, and generation settings.
The full-benchmark comparison additionally sets a Qwen3.8-27B developer against RTC-2B; the same-model comparison (Appendix~\ref{app:same-model}) removes this difference on the 31-task subset.

The GameWorld human reference comes from one novice player under a cap of 100 primitive actions per task~\citep{ouyang2026gameworld}, a protocol that differs from ours; it attains 64.1\% progress, compared with 65.21\% for independent best-of-four, which selects the best of four development runs after the fact.

Each independent session and each collaborative branch has its own submission and output-token allowances.
Peer review consumes the original allowance, and unused budget is not transferred between developers.
Branches that terminate before the review point do not restart; missing reviews do not trigger replacement runs.
These rules preserve each developer's original allowance while permitting different actual expenditures and incomplete exchanges.

The comparison measures attained development performance: the best progress and any successful valid submission across four developers.
Re-evaluating a selected policy on fresh executions would additionally measure its reliability.
The Snake case in Section~\ref{sec:development-budget} and the additional cases in Appendix~\ref{app:development-cases} trace the prior strategy, inspected or received evidence, code changes, and subsequent outcomes, showing how peer evidence enters a revision; the aggregate comparison evaluates which allocation attains higher scores.

For the best-of-$k$ analysis, we enumerate all $\binom{4}{k}$ subsets of the four existing independent sessions for each task.
We take maximum progress and any success within each subset, average over subsets, then average over the 170 tasks, retaining failures as zero.
Thus $k=1$ recovers the independent mean and $k=4$ the per-task Max; intermediate points require no additional development runs.
The curve measures the best performance attained as the number of sessions grows, selected after development by evaluator score.

\subsection{Additional Controller Comparisons}
\begin{table}[ht]
  \centering
  \small\setlength{\tabcolsep}{6pt}\renewcommand{\arraystretch}{1.12}
  \begin{tabular}{lcrr}
\toprule
Controller & $F$ (s) & PG & SR \\
\midrule
GameWorld baseline & \textemdash & 13.83 [9.81, 19.50] & 4.03 [3.23, 6.45] \\
RTC & 3 & 5.08 [4.11, 7.63] & 0.81 [0.00, 3.23] \\
\bottomrule
\end{tabular}

  \caption{Qwen3.5-2B under the GameWorld baseline and RTC. Both use the 31-task subset with four runs per task. Scores are Mean [Min, Max].}
  \label{tab:base2b}
\end{table}

Qwen3.5-2B obtains higher progress through the GameWorld baseline than through RTC.
The two interfaces differ in context and action representation, and the untrained model struggles to produce valid RTC chunks.
Together with the SFT and RL comparisons, this result shows that an execution interface pays off only once the controller has learned to use it.

The full category breakdown for the trained controllers and the Qwen3.8-27B references appears in Table~\ref{tab:rtc-training}.

\subsection{Latency Measurements}
\label{app:latency-measurements}
The timing analysis covers the same 868 episodes as the controller experiments, of which 860 have valid scores, with 38,238 recorded requests across all episodes.
Task scores retain all episodes in their original denominators; the operating statistics use score-valid episodes and report their coverage in Table~\ref{tab:latency-coverage}.
Request latency is the client interval from issuing a request to recording its complete reply, including service queuing, transmission, generation, and client processing.
Official multi-agent requests are tracked separately for each agent; overlapping request durations are not summed into gameplay time.

\begin{table}[ht]
  \centering
  \small\setlength{\tabcolsep}{5pt}\renewcommand{\arraystretch}{1.1}
  \begin{tabular}{@{}lrrrrr@{}}
\toprule
\rowcolor{metricband}
Controller & \shortstack{Valid\\episodes} & Complete & Released & \shortstack{Deadline\\incomplete} & \shortstack{Other\\incomplete} \\
\midrule
Qwen3.8-27B baseline & 123 & 9345 & \textemdash & 79 & 0 \\
Qwen3.8-27B + RTC (3s) & 122 & 4288 & 4166 & 0 & 0 \\
Qwen3.8-27B + RTC (6s) & 123 & 3109 & 2986 & 0 & 0 \\
Qwen3.5-2B + RTC (3s) & 123 & 1257 & 1134 & 0 & 0 \\
RTC-2B (SFT), 3s & 123 & 5986 & 5863 & 0 & 0 \\
RTC-2B (RL), 3s & 122 & 5872 & 5750 & 0 & 0 \\
Qwen3.5-2B baseline & 124 & 8178 & \textemdash & 103 & 0 \\
\bottomrule
\end{tabular}

  \caption{\textbf{Latency sample coverage.} Each condition has 124 episodes on the 31-task subset. Counts here use score-valid episodes. Complete denotes full API replies, including syntactically invalid or late replies; Released is the complete subset whose response gate was not cancelled. Deadline-incomplete and other-incomplete requests are excluded from complete-reply quantiles. The Qwen3.5-2B baseline additionally has request P50/P95 of 1.37/5.80s.}
  \label{tab:latency-coverage}
\end{table}

Effective response latency extends from request start to release after any configured response-floor wait, excluding cancelled gates.
It does not measure the delay from a screenshot to the first physical action, which can include waiting behind committed actions and tick dispatch.
P50/P95 are pooled-request quantiles with linear interpolation; games with more requests contribute more observations.
Complete replies that fail action parsing remain in the request distribution.
Incomplete requests are retained as failures or right-censored observations, with counts reported separately.
Complete replies that arrive too late for gameplay also remain in the request distribution but are excluded from release quantiles when their gates are cancelled.
Consequently, request and release columns use different sample sets: for Qwen3.8-27B RTC at 3s, the full-request median is 3.91s, while the median request duration on the release-eligible subset is 3.89s, followed by a release median of 3.89s.
The lower release median reflects this change of sample set; each individual released reply still waits until its response floor is satisfied.

Syntactic validity is the number of successfully parsed chunks divided by complete replies; a syntactically valid late chunk need not be executed.
The empty-queue fraction is the fraction of dispatched ticks without an assigned queue action.
It includes startup and reset periods but excludes explicitly generated wait actions.
We compute it per valid episode, average within each task, then average across all 31 tasks; missing diagnostics are not assigned zero.
Each RTC condition has diagnostic coverage for every task, with valid-episode counts given above.
Service hardware and load varied across evaluation batches; the latency statistics characterize these observed deployment conditions.

\subsection{Initialization Comparison}
\label{app:initialization}
The initialization comparison uses the 31-task subset, with 124 initial-frame development sessions and the corresponding 124 trajectory-initialized sessions from the full-benchmark study.
The first-frame condition receives the same initial screenshot as the trajectory condition, together with public task instructions and the same tool interfaces, but no initial action records or subsequent rollout frames.
Each condition allows 256K output tokens and at most eight code submissions of up to 60 seconds per development session.
The initial-frame runs use the fixed game seed 42 across repeats; four runs are independent development attempts, not four distinct game seeds.
The two conditions were collected on different dates with different hardware and service load; runs are paired by task and development-run index.

Two initial-frame sessions and eight trajectory-initialized sessions have no valid submission score and remain as zero in the fixed 124-session denominators.
The two initial-frame sessions exhausted their output allowances without producing a valid scored submission.
Restricting to the 115 pairs with valid scores on both sides yields mean PG of 32.72\% and 34.83\%, respectively.
For uncertainty summaries, we resample the 31 task-level paired differences 20,000 times with seed 20261105.
The descriptive 95\% interval for the trajectory-minus-frame difference is $[-4.76,5.12]$ percentage points for mean PG and $[-6.65,12.63]$ for per-task maximum PG.
Both intervals span zero.

Submission counts are available for all 124 initial-frame sessions and 123 trajectory-initialized sessions.
The common 123 pairs contain 735 and 677 submissions, respectively.
Seven trajectory-initialized sessions have zero submissions; restricting to pairs with valid scores reduces the mean-count difference from 0.47 to 0.29 submissions.
A submission is a code-evaluation invocation, not a model request or a fixed amount of gameplay: failed submissions can consume an allowance, and development can continue after reaching the goal.
The trajectory condition additionally uses an initial RTC rollout of up to 60 seconds, shared by the four development sessions for each task.
Its gameplay cost is additional to the reported code-submission counts.

\subsection{Same-Model Comparison}
\label{app:same-model}
\begin{table}[!t]
  \centering
  \footnotesize\setlength{\tabcolsep}{3pt}\renewcommand{\arraystretch}{1.1}
  \begin{tabular}{@{}llrrrr@{}}
\toprule
\rowcolor{metricband}
Code start & Direct control & $\Delta$PG & $\Delta$Max PG & $\Delta$SR & W / T / L \\
\midrule
Trajectory & Qwen3.8-27B + RTC (6s) & \shortstack[r]{+5.84\\[-1pt]{\scriptsize\color{black!65}[$-$1.60, +13.15]}} & \shortstack[r]{+13.45\\[-1pt]{\scriptsize\color{black!65}[+1.55, +25.81]}} & \shortstack[r]{+3.23\\[-1pt]{\scriptsize\color{black!65}[$-$3.23, +9.68]}} & 20 / 3 / 8 \\
Trajectory & Qwen3.8-27B + RTC (3s) & \shortstack[r]{+10.47\\[-1pt]{\scriptsize\color{black!65}[+3.59, +17.34]}} & \shortstack[r]{+17.83\\[-1pt]{\scriptsize\color{black!65}[+5.64, +30.29]}} & \shortstack[r]{+8.87\\[-1pt]{\scriptsize\color{black!65}[+1.61, +17.74]}} & 21 / 5 / 5 \\
Trajectory & Qwen3.8-27B, GameWorld baseline & \shortstack[r]{+7.31\\[-1pt]{\scriptsize\color{black!65}[$-$0.22, +15.35]}} & \shortstack[r]{+18.16\\[-1pt]{\scriptsize\color{black!65}[+7.54, +30.13]}} & \shortstack[r]{+7.26\\[-1pt]{\scriptsize\color{black!65}[+0.00, +16.13]}} & 17 / 3 / 11 \\
Trajectory & RTC-2B (RL), direct (3s) & \shortstack[r]{+10.48\\[-1pt]{\scriptsize\color{black!65}[+1.60, +19.80]}} & \shortstack[r]{+17.85\\[-1pt]{\scriptsize\color{black!65}[+5.44, +30.51]}} & \shortstack[r]{+7.26\\[-1pt]{\scriptsize\color{black!65}[$-$1.61, +16.94]}} & 20 / 4 / 7 \\
\midrule
Frame & Qwen3.8-27B + RTC (6s) & \shortstack[r]{+5.80\\[-1pt]{\scriptsize\color{black!65}[+0.11, +11.56]}} & \shortstack[r]{+11.11\\[-1pt]{\scriptsize\color{black!65}[+1.03, +20.79]}} & \shortstack[r]{+0.00\\[-1pt]{\scriptsize\color{black!65}[$-$4.03, +4.03]}} & 22 / 3 / 6 \\
Frame & Qwen3.8-27B + RTC (3s) & \shortstack[r]{+10.44\\[-1pt]{\scriptsize\color{black!65}[+3.73, +17.93]}} & \shortstack[r]{+15.49\\[-1pt]{\scriptsize\color{black!65}[+5.18, +26.18]}} & \shortstack[r]{+5.65\\[-1pt]{\scriptsize\color{black!65}[$-$1.61, +14.52]}} & 24 / 2 / 5 \\
Frame & Qwen3.8-27B, GameWorld baseline & \shortstack[r]{+7.27\\[-1pt]{\scriptsize\color{black!65}[+0.56, +15.28]}} & \shortstack[r]{+15.82\\[-1pt]{\scriptsize\color{black!65}[+7.02, +25.49]}} & \shortstack[r]{+4.03\\[-1pt]{\scriptsize\color{black!65}[$-$4.03, +13.71]}} & 17 / 5 / 9 \\
Frame & RTC-2B (RL), direct (3s) & \shortstack[r]{+10.44\\[-1pt]{\scriptsize\color{black!65}[+2.32, +19.65]}} & \shortstack[r]{+15.51\\[-1pt]{\scriptsize\color{black!65}[+4.91, +26.63]}} & \shortstack[r]{+4.03\\[-1pt]{\scriptsize\color{black!65}[$-$4.03, +13.71]}} & 22 / 3 / 6 \\
\bottomrule
\end{tabular}

  \caption{\textbf{Paired comparison of code development and direct control on the 31-task subset.} Entries are code minus direct, in percentage points, for task-averaged mean PG, per-task maximum PG, and mean SR, with descriptive 95\% bootstrap intervals; W / T / L counts tasks where code development has higher, equal, or lower mean progress.}
  \label{tab:same-model-paired}
\end{table}

Table~\ref{tab:same-model-paired} pairs each code-development condition with each direct-control configuration on the 31-task subset, with four runs per task on both sides.
Intervals resample the 31 tasks 20,000 times with seed 20261107; because each task comes from a different game, this is a game-level bootstrap.
The intervals are descriptive: they reflect task sampling for fixed runs and are uncorrected for multiple comparisons.
The direct-control and code-development runs were collected in different batches, with different dates, hardware, and service load.
A code-development run reports its best valid submission and a direct-control run one continuous play, so Max PG uses the same aggregation over a different number of executions per run.
For mean PG, the interval includes zero for the trajectory start against the 6s configuration, $[-1.60,13.15]$, and against the GameWorld baseline; every per-task maximum PG interval excludes zero.
Mean SR differences against the 6s configuration are small, with intervals $[-3.23,9.68]$ for the trajectory start and $[-4.03,4.03]$ for the single-frame start.
The subset uses only the \texttt{05} objectives, while the full benchmark includes all five tasks per game; their aggregate margins therefore describe different task distributions.

\subsection{Task- and Game-Level Comparisons}
\label{app:task-game-level}
The main text reports progress and success by category; this section resolves the same comparisons by task and by game.

\paragraph{RTC-2B and Qwen3.8-27B direct control.}
On the 31-task subset, RTC-2B is ahead of Qwen3.8-27B at 3s on 11 tasks, tied on 10, and behind on 10 in four-run mean progress; against Qwen3.8-27B at 6s, it is ahead on 13, tied on six, and behind on 12.
Within Puzzle, it reaches 51.29\% on Tetris, above Qwen3.8-27B at 3s (46.33\%) and 6s (39.75\%), and 36.38\% on Hextris, above Qwen3.8-27B at 3s (30.97\%).
The category gap comes mainly from Wordle, 5.00\% against 100\% for both Qwen3.8-27B configurations, and 2048, 21.29\% against 54.30\% and 91.41\%, with a smaller gap on the Astray maze.

\paragraph{Code development and direct RTC-2B control on 170 tasks.}
\begin{table}[!ht]
  \centering
  \small\setlength{\tabcolsep}{5pt}\renewcommand{\arraystretch}{1.12}
  \begin{tabular}{@{}lrrrr@{}}
\toprule
\rowcolor{metricband}
Progress per task & Code higher & Tie & RTC higher &  \\
\midrule
Mean of four runs & 111 & 31 & 28 &  \\
Best of four runs & 91 & 65 & 14 &  \\
\midrule
\rowcolor{metricband}
Solved in any of four runs & Code only & Both & RTC only & Neither \\
\midrule
Number of tasks & 41 & 35 & 5 & 89 \\
\bottomrule
\end{tabular}

  \caption{\textbf{Task-level comparison of independent code development and direct RTC-2B control.} Top: number of the 170 tasks on which each side has higher progress, using four-run means and four-run bests. Bottom: number of tasks solved in at least one of four runs. Both conditions score the five Captain Callisto tasks as zero.}
  \label{tab:task-level}
\end{table}

\begin{figure}[!ht]
  \centering
  \includegraphics[width=.92\linewidth]{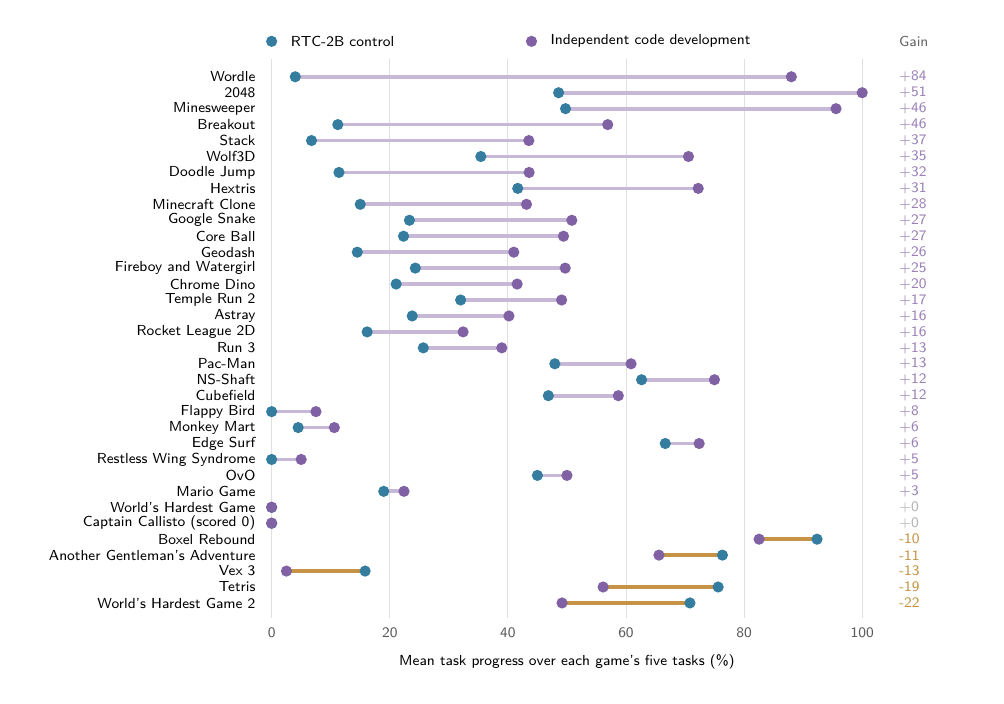}
  \caption{\textbf{Code development improves progress on most games.} Mean task progress over each game's five tasks, each averaged over four runs, for direct RTC-2B control and independent code development, sorted by gain in points; gold segments mark the five games where direct control is higher. Captain Callisto is scored zero for both.}
  \label{fig:game-gains}
\end{figure}

Four-run mean progress is higher for code development on 111 of the 170 tasks, higher for direct control on 28, and tied on 31, including the five Captain Callisto tasks that both conditions score as zero (Table~\ref{tab:task-level}).
Comparing the best of four runs on each side, code development is higher on 91 tasks, direct control on 14, and 65 are tied.
Across games, code development has higher mean progress on 27 of 34 (Figure~\ref{fig:game-gains}), led by Wordle (+84 points), 2048 (+51), Minesweeper (+46), and Breakout (+46).
Direct control keeps the advantage on five games, most clearly World's Hardest Game 2 ($-22$ points) and Tetris ($-19$), and both conditions score zero on World's Hardest Game and Captain Callisto.
In success, 76 tasks are solved in at least one of four code runs and 40 in at least one of four direct runs; 41 are solved only by code development, five only by direct control, and 89 by neither.

\paragraph{Same-model and team comparisons.}
With the model fixed on the 31-task subset, code development has higher mean progress than Qwen3.8-27B RTC at 6s on 20 tasks from a trajectory and on 22 from a single frame, and lower on 8 and 6 (Table~\ref{tab:same-model-paired}).
On the 170 tasks, the collaborative team's best progress is higher than the independent best of four on 27 tasks, equal on 95, and lower on 48.

\FloatBarrier
\section{Training RTC-2B}
\label{app:rtc-training}

This appendix describes how RTC-2B (RL), evaluated in Sections~\ref{sec:exp-training} and~\ref{sec:exp-si}, was obtained.
Training proceeds in three stages: we prepare a separate collection of browser games with host-side score access, distill RTC behavior from Qwen3.8-27B through SFT, and continue with online RL on the same games.
All training precedes the GameWorld evaluation, which uses different games, a continuously running clock, and the official task scoring.

\subsection{Training Games}
\label{app:training-games}

The training environment is built from a source snapshot of 2,522 browser games from js13kgames~\citep{js13kgames}, not from GameWorld tasks.
A coding agent analyzes each game's source.
It reports the objective, controls, start and end conditions, a canonical numeric score reference with source evidence for its initialization and updates, and the location at which that score can be read.
The score must be a value the game itself maintains; constructed expressions, proxy metrics, and partial sums are not accepted.

A deterministic filter then applies two groups of conditions (Table~\ref{tab:training-games}).
The runtime conditions reject games with missing, external, or out-of-tree resources, backend interfaces, media input, or repository-specific runtimes; gamepad, sensor, and WebXR dependencies are sent to review.
The reward conditions require a high-confidence score that increases with performance, changes continuously or frequently, depends on player control, does not accumulate passively, and is suitable for differencing.
The runtime checks are programmatic, whereas the reward properties come from the agent's static source analysis.

\begin{table}[h]
  \centering
  \small\setlength{\tabcolsep}{6pt}\renewcommand{\arraystretch}{1.1}
  \begin{tabular}{@{}llr@{}}
    \toprule
    \rowcolor{metricband}
    Stage & Outcome & Games \\
    \midrule
    Source snapshot & Discovered & 2,522 \\
    \midrule
    Score analysis & Eligible / rejected & 853 / 290 \\
     & No score found / uncertain / failed & 923 / 167 / 289 \\
    \midrule
    Strict filter & Pass / reject / review & 395 / 1,667 / 460 \\
    \midrule
    Instrumentation & Games with a host score reader & 395 \\
    \bottomrule
  \end{tabular}
  \caption{\textbf{Construction of the training-game collection.} The analysis outcomes are those of the static source analysis; only games that pass the strict filter enter training.}
  \label{tab:training-games}
\end{table}

For each of the 395 games, a copy of the game receives one inserted line at the analyzed location, for example \texttt{window.\_\_RL\_GET\_SCORE\_\_ = () => E;} where \texttt{E} is that game's score variable.
The inserted reader does not change gameplay logic and is visible only to the host, which uses it for training rewards.
The policy never receives the source, the analysis, or the score reader; a score drawn on screen remains part of its RGB observation.

The training and evaluation collections use distinct game implementations and share several game families, including Snake, Tetris, Breakout, Flappy Bird, and dinosaur runners.
The evaluation measures transfer to new implementations and tasks within these familiar families as well as other games.

\subsection{RL Interaction Protocol}
\label{app:training-protocol}

RL rollouts use the RTC action format of Section~\ref{sec:rtc}.
A tick lasts 200\,ms and may press one to four keys, click once with the left button at normalized coordinates, or both; every input is released at the end of the tick.
Each request is built afresh from up to seven frames at the offsets of Appendix~\ref{app:rtc-timing}, the run-length-compressed actions executed between them, the committed queue, the request index, the remaining budget, the first tick available for new actions, the required number of lines, and any previous syntax error.
Frames are captured at $1280\times720$ and sent as JPEG at quality 80.
The system message states the action protocol and a single generic goal: discover the game from the images, start it, make progress, and recover or reset after failure.
Unlike GameWorld evaluation, RL rollouts provide no per-game instructions, controls, or objectives.
The host does not click through menus, and episodes end when the request or tick budget is exhausted, because the 395 games have no uniformly reliable end-of-game signal.

RL rollouts use a virtual clock.
The browser is paused while the model generates; after the reply arrives, the host advances 15 committed ticks before appending the new chunk.
New actions therefore always act on an observation at least 3\,seconds old, independently of service queueing, weight synchronization, or actual generation time.
Each episode starts with 15 neutral ticks, and the scheduler asks the model to fill a 30-action planning capacity, so steady-state requests usually require 15 lines.
A reply with the wrong number of lines or any invalid action is rejected as a whole; it still consumes a request, the existing queue continues, and neutral ticks fill any gap.
A reset reloads the same game without restoring the budget, and replies based on the previous session are discarded.
RL episodes allow at most 32 model requests and 960 ticks.

This protocol fixes the delay that the policy must anticipate while keeping training independent of cluster timing.
It differs from evaluation, where the browser keeps running and the delay is the actual request time subject to the response floor (Table~\ref{tab:train-eval}).

\subsection{Supervised Fine-Tuning}
\label{app:training-sft}

The SFT data are RTC episodes of Qwen3.8-27B on the 395 training games, using the mixed-history context with at most 100 requests per episode.
From 382 usable episode archives containing 33,737 requests, we keep the 31,626 requests whose replies are complete and whose action chunks are syntactically valid.
Requests are not filtered by score or success: 2,852 of the retained requests come from episodes that ended with a harness failure but are themselves complete and valid.
SFT therefore produces RTC-2B (SFT), which learns to reproduce Qwen3.8-27B's RTC behavior on the training games, rather than imitating only successful play.

Each example supervises the final assistant reply, including its reasoning and action chunk, while the request context and images serve as input with zero loss.
Examples are rendered with the model's native chat template with thinking enabled.
Rendered examples have a median of 8,415 and a maximum of 17,121 tokens, with 8,691,912 supervised tokens in total.
Starting from Qwen3.5-2B, we update all 2,213,241,664 parameters, including the vision encoder, projector, and language model, without LoRA.
Training uses LlamaFactory with FSDP2 on four GPUs, BF16, and activation checkpointing; one example per GPU with two-step gradient accumulation gives a global batch of eight.
We optimize with AdamW at a learning rate of $10^{-5}$ for one epoch with seed 42, for 3,953 updates.

\subsection{Online Reinforcement Learning}
\label{app:training-rl}

RL starts from the SFT checkpoint and samples episodes online on the same 395 games under the protocol above.
Each update draws four games and runs eight independent episodes per game, for 32 nominal episodes; the number of requests that contribute to the loss depends on the filtering described below.
The request-level variant of GRPO described below computes a separate advantage for each reply, with no critic, no reference model, and a zero KL coefficient.

\paragraph{Request rewards.}
Each executed tick records the request whose reply produced it, and the host reads the game score at tick boundaries.
For request $j$ with executed interval $[t^{\mathrm{s}}_j, t^{\mathrm{e}}_j]$, the reward is the score change over the ticks it actually controlled,
\begin{equation}
  r_j = s\bigl(t^{\mathrm{e}}_j\bigr) - s\bigl(t^{\mathrm{s}}_j\bigr).
\end{equation}
The interval follows executed actions rather than the time between request and reply, so earlier committed actions remain attributed to earlier requests, and neutral filler ticks are attributed to no request.
Requests are masked when they span a reset, are executed only partially, lack a reliable score reading at either boundary, have unclear attribution, or have a negative delta.
Valid zero deltas are kept.
Masking every negative delta is a conservative choice, because the score reader cannot distinguish a game penalty from a hidden reset; the controller therefore receives no direct signal from score losses.

\paragraph{Returns and group-relative advantages.}
Within each maximal run of consecutive valid requests in an episode, the return of request $j$ is the undiscounted sum of the rewards from $j$ to the end of that run, $G_j=\sum_{k\ge j} r_k$; a masked request or an execution gap ends the run.
For the eight episodes of a game, requests are aligned by their index in the episode.
This aligns their order of interaction; the episodes can visit different game states.
The baseline for episode $e$ at index $j$ averages the returns of the other episodes that have a valid request at that index, and the advantage divides the residual by a scale $\sigma$ shared within the game group:
\begin{equation}
  A_{e,j} = \frac{G_{e,j} - \frac{1}{|V_j\setminus\{e\}|}\sum_{e'\in V_j\setminus\{e\}} G_{e',j}}{\sigma},
\end{equation}
where $V_j$ is the set of episodes with a valid request at index $j$, and $\sigma$ is the population standard deviation of the residuals after centering each index.
Requests without a valid peer do not enter the policy loss, a zero scale gives zero advantage, and a game contributes only when at least four of its episodes are usable.
Each request's advantage is assigned to every token of its reply, so longer replies contribute more tokens to the loss.

\paragraph{Training procedure.}
Initial experiments with episode-level GRPO showed no clear improvement, so we adopted the request-level objective above.
The resulting checkpoint is RTC-2B (RL).
Rollouts sample with temperature 0.6 and top-$p$ 0.95 with thinking enabled, a 65,536-token limit per request, and at most 8,192 output tokens.
The policy loss uses PPO-style ratio clipping with $\epsilon=0.4$ and recomputed log-probabilities, and allows sampled data to lag the trained policy by up to three versions.
We use Adam with a constant learning rate of $5\times10^{-7}$, weight decay 0.01, and gradient clipping at 1.0.
Training runs on eight GPUs on one node, four for the AReaL/Megatron actor (two-way tensor and two-way context parallelism) and four for SGLang rollout replicas.

\subsection{Training and Evaluation Conditions}
\label{app:train-eval}

Table~\ref{tab:train-eval} lists the differences between RL training and the evaluation in the main text.
The evaluation therefore tests transfer from a fixed simulated delay on generic, uninstructed games to continuously running GameWorld tasks with instructions and actual latency.
Section~\ref{sec:exp-training} compares the original model, the SFT checkpoint, and the final RTC-2B (RL) checkpoint.

\begin{table}[h]
  \centering
  \footnotesize\setlength{\tabcolsep}{4pt}\renewcommand{\arraystretch}{1.12}
  \begin{tabular}{@{}p{0.2\linewidth}p{0.37\linewidth}p{0.37\linewidth}@{}}
    \toprule
    \rowcolor{metricband}
    & RL training & GameWorld evaluation \\
    \midrule
    Games & 395 filtered js13kgames~\citep{js13kgames} & 31-task subset or full 170-task benchmark \\
    Instructions & One generic goal for all games & Official task goal and controls \\
    Game clock & Virtual; paused during inference & Continuously running \\
    Delay & Fixed simulated 3\,s & Actual request time, with a 3\,s floor \\
    Budget & 32 requests, at most 960 ticks & 180\,s per run \\
    Policy input & RGB frames, executed actions, committed queue & Same, with task instructions \\
    Signal & Host score reader; request-level score change & Official PG and SR \\
    \bottomrule
  \end{tabular}
  \caption{\textbf{RL training and evaluation conditions of RTC-2B.} Both use a 200\,ms tick and queue-conditioned visual context. The host score reader is used only for training rewards. SFT data come from Qwen3.8-27B episodes on the same 395 games with up to 100 requests.}
  \label{tab:train-eval}
\end{table}

\FloatBarrier
\section{Specializing the Controller's Weights: Stack}
\label{app:training-si}

This appendix details the weight-level specialization summarized in Section~\ref{sec:exp-training}.
Starting from RTC-2B (RL), GPT-6 Astra~\citep{openai2026astra} led an effort to improve it on a single GameWorld task, Stack \texttt{27\_05}, whose goal is a 30-layer tower.
Astra constructed an independent simulator, designed the training data and procedures, called Qwen3.8-27B to generate complete reasoning and actions as supervision, trained and evaluated the controller, and revised the procedure from the results; humans adjusted the effort only between stages.

\subsection{Original-Game Evaluation}
\label{app:stack-original}
We evaluate RTC-2B and its specialized SFT and RL checkpoints in the original GameWorld game, continuously running, with a 180-second window, 200\,ms tick, 15-line chunks, and 3-second response floor.
Each controller uses seeds 42--45, with thinking enabled, temperature 0.6, and up to 8,192 output tokens.
The RTC-2B reference reuses the Stack trials from the main evaluation; the specialized SFT and RL checkpoints share an inference service, harness, and reward.

\begin{table}[ht]
  \centering
  \small\setlength{\tabcolsep}{5pt}\renewcommand{\arraystretch}{1.12}
  \begin{tabular}{lcrr}
\toprule
Controller & Heights & PG (\%) & SR (\%) \\
\midrule
RTC-2B & 2, 1, 3, 1 & 5.83 [3.33, 10.00] & 0.00 [0.00, 0.00] \\
+ Stack SFT & 1, 3, 3, 2 & 7.50 [3.33, 10.00] & 0.00 [0.00, 0.00] \\
+ Stack RL & 2, 4, 5, 2 & 10.83 [6.67, 16.67] & 0.00 [0.00, 0.00] \\
\bottomrule
\end{tabular}

  \caption{\textbf{Stack specialization in the original game.} Heights are the highest layers reached in four 180-second games, ordered by seed 42--45. PG is height divided by 30, shown as Mean [Min, Max]; no run completes the tower. The RL checkpoint uses the audited reward.}
  \label{tab:stack-original}
\end{table}

RTC-2B reaches a mean height of 1.75 layers (5.83\% progress), compared with 2.25 layers (7.50\%) for specialized SFT.
RL raises the mean height by one layer, from 2.25 to 3.25 of 30, and task progress from 7.50\% to 10.83\%.
RL improves on its SFT parent on three seeds and ties on one; the best game reaches five layers, and no controller completes the tower.
These are end-to-end scores that include actual inference latency: specialized SFT and RL have median request times of 3.37\,s and 4.15\,s, with empty queues on 44.97\% and 41.78\% of ticks, respectively.

\subsection{Simulator and Interface}
\label{app:stack-simulator}
Training uses Astra's independent pygame implementation of Stack's core mechanics: a block that slides along alternating axes and bounces off the edges, a space key that drops it, trimming of the overhang, failure at zero overlap, and a 30-layer goal.
The simulator uses no assets, levels, or trajectories from the original game.
Initial block widths are drawn from 170 to 220 units and speeds from 65 to 115 units per second, and drops within two units of perfect alignment snap into place.
The local quality of a drop is the fraction of the block's area that survives it; small errors compound, since a 95\% drop repeated 30 times leaves about 21.5\% of the original area.

The controller uses the same RTC request format as the main evaluation: up to seven frames at the offsets of Appendix~\ref{app:rtc-timing}, the executed inputs, the committed queue, and timing fields, and it returns 15 one-tick actions.
It never sees the block's coordinates, speed, the reference action, or the reward.
The simulator pauses during generation, then advances 3 seconds while executing committed actions before appending the new chunk.
Any remaining queued actions execute before that chunk, whose first action is dispatched no earlier than the next tick.
A full evaluation game lasts 900 ticks, or 180 seconds of simulated time, and restarts do not restore the budget.

\subsection{Supervised Specialization}
\label{app:stack-sft}
Specialization began with SFT on data generated in the simulator.
The Qwen3.8-27B teacher receives private timing hints while writing its reasoning and actions; the hints are removed from the student's input, and the teacher's complete reasoning and action chunk serve as targets.
Astra then explored a range of procedures starting from RTC-2B, each evaluated over eight simulated games (Table~\ref{tab:stack-sft-explore}).
Full-parameter SFT on 512 mixed examples, followed by 36 examples with shorter complete reasoning, produced the short-reasoning checkpoint; continuing on 206 examples that vary block speed and scene layout with full-parameter training for 1,024 steps (learning rate $10^{-6}$, batch 4) produced the visual-clock checkpoint, denoted RTC-2B + Stack SFT.
Preference optimization, rejection-sampling fine-tuning, correction data, larger data mixtures, and an early progress-based RL run did not improve on these checkpoints.

\begin{table}[ht]
  \centering
  \footnotesize\setlength{\tabcolsep}{5pt}\renewcommand{\arraystretch}{1.1}
  \begin{tabular}{@{}lr@{}}
    \toprule
    \rowcolor{metricband}
    Adaptation of RTC-2B (complete reasoning) & Mean height \\
    \midrule
    RTC-2B (RL) & 2.00 \\
    Full SFT, 512 mixed examples & 3.63 \\
    \quad + 36 short-reasoning examples (short-reasoning checkpoint) & 4.88 \\
    \quad\quad + 206 speed and layout examples, 1,024 steps (visual-clock checkpoint) & 4.88 \\
    Full SFT + 83 response pairs, DPO & 3.88 \\
    Short-reasoning + 97 response pairs, DPO & 4.38 \\
    Correction data, 241 long or 192 short examples & 3.88 / 3.75 \\
    189 examples from four successful worlds, 1,024 steps & 4.50 \\
    Vision module only, 206 examples & 3.38 \\
    \bottomrule
  \end{tabular}
  \caption{\textbf{Supervised procedures explored in the simulator.} Mean of the highest of 30 layers over eight simulated games on a fixed seed group. On a second seed group with varied layouts, the visual-clock checkpoint reaches 5.88 layers against 3.63 for its parent. The procedures differ in data and budget and record Astra's exploration.}
  \label{tab:stack-sft-explore}
\end{table}

\subsection{Reinforcement Learning in a Simulator}
\label{app:stack-rl}
Using the simulator to sample additional training interactions, Astra next trained the visual-clock checkpoint with request-level RL.
Each update fixes a simulator state and its public RTC context, samples eight complete replies (four sessions with two candidates each), executes each on its own copy of the world, and scores the result.
States come from a curriculum that warms up the tower with a reference controller to a maximum starting height of 3, 8, 16, and then 29 layers, over 16, 16, 32, and 64 updates; one in four states calls for waiting and three in four offer a drop.
Training uses rank-8 LoRA on the language model's attention projections, Adam with a learning rate of $5\times10^{-5}$, PPO-style clipping at 0.4, no critic or KL penalty, temperature 1.0, and up to 8,192 output tokens with complete reasoning.
Each run comprises 128 updates and 1,024 candidate replies.
Before these runs, Astra had tried whole-episode returns, same-state branching, and a low-height progress reward over up to 32 updates each; none improved full-game height.

\paragraph{Relative-advantage run.}
The first run rewards a safe drop with $0.2+0.8q^4$, where $q$ is the drop quality, penalizes failures and missed opportunities, and normalizes rewards within each group of eight.
An audit of all 1,024 candidates found that 115 safe drops keeping less than half the block still received positive advantage, because a poor drop can beat its group.
In one case, a drop keeping 0.89\% of the block received an advantage of $+1.59$; handing the game to the reference controller after this drop ends at two layers, whereas waiting or following the reference completes the tower.

\paragraph{Audited-reward run.}
The second run therefore restarts from the same parent with a stricter objective.
A drop earns positive reward only when its quality reaches $\max(0.90, q_{\mathrm{ref}}-0.01)$, where $q_{\mathrm{ref}}$ is the quality of the reference action, and the advantage equals the reward itself rather than its deviation from the group mean, so a poor drop never receives positive advantage.
Re-scoring the first run's candidates confirmed that all 115 low-quality positives become negative.
During training, no below-threshold drop received positive advantage, but only 76 of 1,024 candidates were precise drops.

\paragraph{Results.}
On eight held-out simulated seeds, the parent reaches 3.50 layers, the relative-advantage run 3.75, and the audited-reward run 4.00, and none completes the tower (Table~\ref{tab:stack-weight-si}).
The same audited-reward adapter is the one evaluated in the original game in Appendix~\ref{app:stack-original}, where it reaches 3.25 layers.
The audited-reward controller remains active rather than learning to wait: it attempts 156 drops against 168 for the parent, with mean surviving area 54.66\% against 48.79\%.
Its precision stays far below what the tower requires.

\paragraph{The interface supports a winning policy.}
Two hand-written reference controllers play 16 new seeds to show what the simulator and interface allow.
A controller reading the true simulator state completes the tower in 16 of 16 games under a fixed 3-second delay and also under latency sampled from request times measured for RTC-2B + Stack SFT.
A controller that receives only the same RGB requests as the student, but knows the simulator's colors and geometry, completes 14 of 16 under a fixed delay and 13 of 16 under sampled latency.
A visual control program can therefore win through this interface, which the trained weights do not reach.

\paragraph{Comparison with code development.}
On the same game, code development averages 43.55\% progress across Stack's five tasks, against 6.75\% for RTC-2B, using at most eight 60-second executions of the original game per run.
On task \texttt{27\_05} itself, code development averages 16.67\% and its best of four runs 36.67\%, about five and eleven layers, against 10.83\% (3.25 layers) for the audited-reward controller in the original game.
For context, the original Qwen3.5-2B model with GameWorld's official interface also reaches 16.67\% on this task, using the baseline execution protocol of Appendix~\ref{app:execution-protocols}.

\FloatBarrier
\section{An Interpretable Code Policy for 2048}
\label{app:2048-code-policy}

A generated 2048 policy completes the task's 1024-point objective using visual parsing and a compact action rule.
The unmodified, 153-line Python program estimates the board from pixels, retains its own state, and selects keyboard actions without runtime calls to a generative model or neural vision service.
We trace its observation-to-action computation and recorded execution to show how task-specific code operates within the live game loop.

\subsection{From Pixels to a Decision Rule}

The policy implements
\begin{equation}
  (a_t,m_{t+1})=\pi(o_t,m_t),
\end{equation}
as in Equation~\ref{eq:code-policy}, where $o_t$ is an RGB screenshot and $m_t$ is the program's retained state.
For each of the 16 board cells, it takes the median RGB value of a $3\times3$ patch near the cell's upper-left corner and matches it to a stored tile palette.
Cell positions and colors were measured from public initial frames.
The program reconciles the observation with the predicted effect of its previous move to handle unknown cells and potentially stale images.
The resulting board is a visual estimate built from pixels alone.

\begin{figure}[!h]
  \centering
  \includegraphics[width=\linewidth]{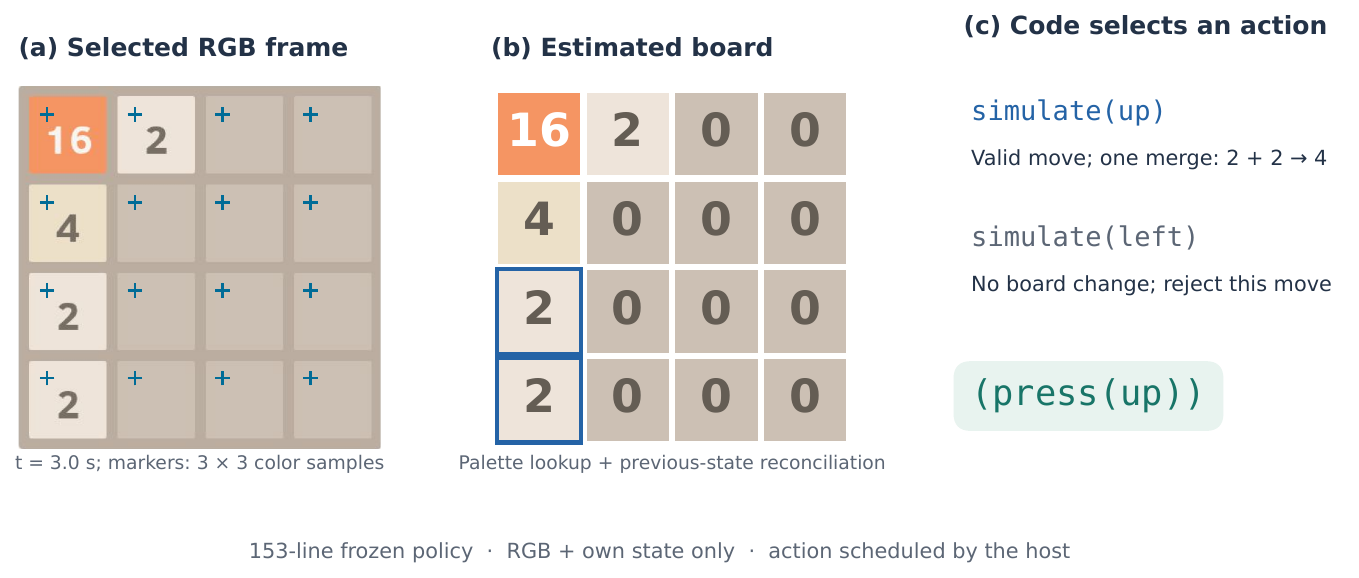}
  \caption{\textbf{A visual observation becomes an action.} The recorded frame was selected at $t=3.0$\,s. Crosses mark the color-sampling patches. Given its estimated board, the policy predicts one merge from moving up, while moving left leaves the board unchanged; it emits \texttt{(press(up))}. Frame selection time is distinct from capture time.}
  \label{fig:2048-rgb-action}
\end{figure}

The decision rule favors up and left, selecting the direction with more merges when both are valid and alternating on ties.
If neither is valid, it checks fallback directions; if no move changes the board, it returns a wait action.
Listing~\ref{lst:2048-choose} shows the rule exactly as generated.
The combination of a task-specific visual representation, a local dynamics model, and a short action-selection routine is sufficient for the recorded task completion.

This program was a developer's first submission in a collaborative team, before peer review, and already covers the complete path from pixels to actions.
Its fixed geometry and palette specialize it to this game rendering, and its only runtime library is NumPy.

\FloatBarrier
\subsection{The Original Policy Interface and Decision Rule}

The following excerpts are reproduced verbatim from the 153-line program.
Together they show how the policy retains its own state and converts successive screenshots into actions; helper functions implement visual parsing, state reconciliation, and deterministic move simulation.

\begin{lstlisting}[language=Python,basicstyle=\ttfamily\small,
  keywordstyle=\color{labblue},commentstyle=\color{black!55},
  columns=fullflexible,breaklines=true,showstringspaces=false,
  lineskip=-1pt,frame=single,rulecolor=\color{black!20},aboveskip=8pt,belowskip=6pt,
  caption={The original direction-selection rule. The helper \texttt{simulate} returns the moved board and merge count. A move is valid when it changes the estimated board.},
  label={lst:2048-choose}]
def choose_move(b, last_move):
    # Prefer up/left (top-left corner strategy)
    up, mu = simulate(b, 'up')
    le, ml = simulate(b, 'left')
    up_ok = not np.array_equal(up, b)
    le_ok = not np.array_equal(le, b)
    if up_ok and le_ok:
        if mu > ml:
            return 'up'
        if ml > mu:
            return 'left'
        # equal: alternate away from last move
        return 'left' if last_move == 'up' else 'up'
    if up_ok:
        return 'up'
    if le_ok:
        return 'left'
    for d in ('down', 'right', 'up', 'left'):
        nb, _ = simulate(b, d)
        if not np.array_equal(nb, b):
            return d
    return None
\end{lstlisting}

\begin{lstlisting}[language=Python,basicstyle=\ttfamily\small,
  keywordstyle=\color{labblue},commentstyle=\color{black!55},
  columns=fullflexible,breaklines=true,showstringspaces=false,
  lineskip=-1pt,frame=single,rulecolor=\color{black!20},aboveskip=8pt,belowskip=6pt,
  caption={The original policy interface. Helper functions and the palette are defined in the complete 153-line program. Persistent state belongs to the policy; no internal game-state API is consulted.},
  label={lst:2048-step}]
def step(frame, state):
    if state is None:
        state = {'board': np.zeros((4, 4), int), 'last_move': None, 'n': 0, 'over': False}
    obs = perceive(frame)
    known = int((obs >= 0).sum())
    if known > 0:
        board = reconcile(obs, state['board'], state['last_move'])
    else:
        board = state['board']
    # detect official reset: non-empty remembered board but empty observed
    if known >= 3 and int((obs > 0).sum()) == 0 and int((state['board'] > 0).sum()) >= 4:
        state['board'] = np.zeros((4, 4), int)
        state['last_move'] = None
        board = np.zeros((4, 4), int)
    else:
        state['board'] = board
    if state['over'] and int((state['board'] > 0).sum()) < 16:
        state['over'] = False
    d = choose_move(board, state['last_move'])
    state['n'] += 1
    if d is None:
        state['over'] = True
        return '(wait())', state
    state['last_move'] = d
    return '(press(%s))' % d, state
\end{lstlisting}

\FloatBarrier
\subsection{Recorded Gameplay and Execution Timing}

\begin{figure}[!h]
  \centering
  \includegraphics[width=.68\linewidth,trim=0 22 0 0,clip]{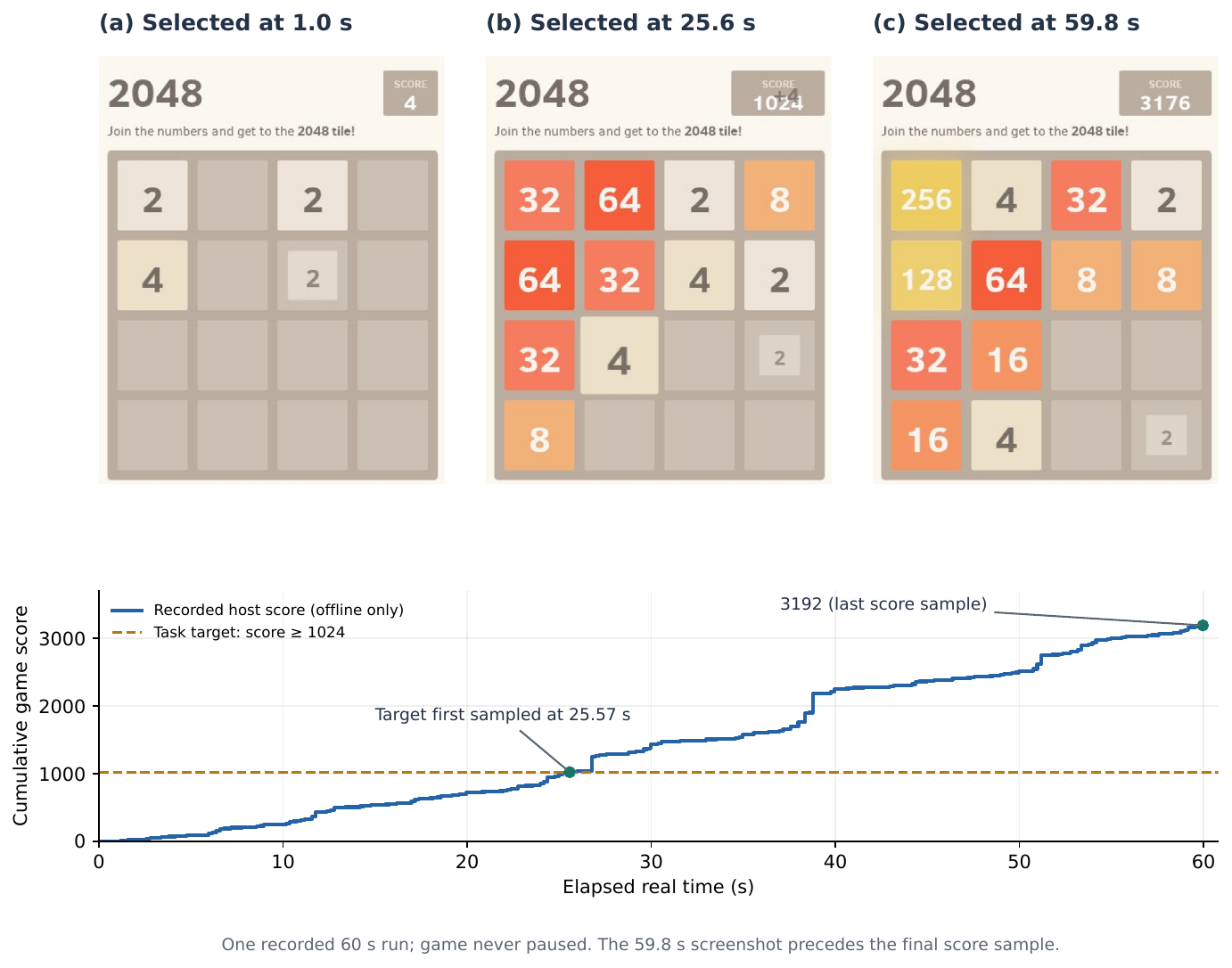}
  \caption{\textbf{A generated policy completes the score objective.} Screenshots and independently sampled host scores from the same episode. The target is first sampled at 25.57\,s. The 59.8\,s screenshot displays 3176 and precedes the final score sample of 3192. Structured score samples were used for reporting, not as policy inputs.}
  \label{fig:2048-episode}
\end{figure}

The task requires a cumulative game score of at least 1024.
In the recorded, non-paused 60-second episode, the target is first sampled at 25.57\,s and the final sampled score is 3192 (Figure~\ref{fig:2048-episode}).
This episode is drawn from the collaborative evaluation.

\begin{figure}[!h]
  \centering
  \includegraphics[width=.85\linewidth,trim=0 0 0 22,clip]{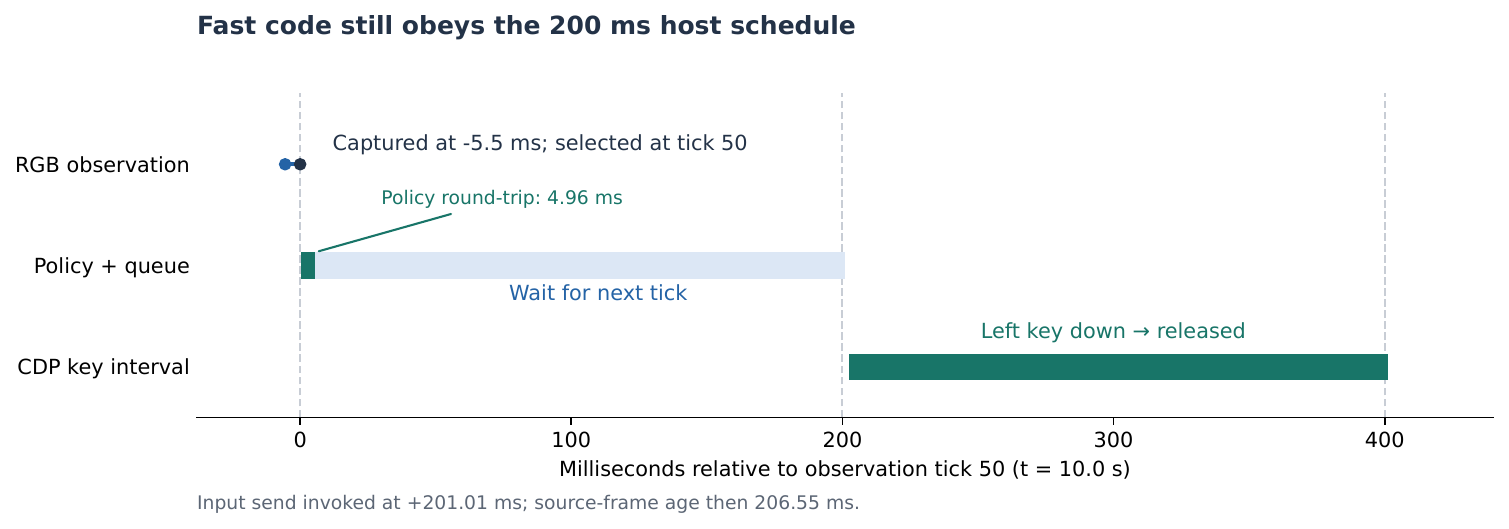}
  \caption{\textbf{Fast computation retains the runtime's execution schedule.} For tick 50 ($t=10.0$\,s), input send is invoked at $t=10.201$\,s, when the source frame is 206.55\,ms old. The green interval spans key-down and release acknowledgements. Observation-to-dispatch delay adds the wait for the next tick to the policy's round trip.}
  \label{fig:2048-timing}
\end{figure}

Median host-measured policy round-trip time is 5.47\,ms (P95: 6.83\,ms), including process communication.
This is distinct from action latency: in Figure~\ref{fig:2048-timing}, computation takes 4.96\,ms, but input send occurs on the next tick, 201.01\,ms after the observation tick.
The record contains 300 policy responses and 298 keyboard-input windows; the initial default wait, one policy wait, and a final queued action account for the difference.

\FloatBarrier

\FloatBarrier
\section{RTC Timing Analysis and Context Construction}
\label{app:rtc-timing}

Queued execution must cover inference while leaving time to respond to new evidence.
We derive the corresponding planning and response constraints under an ideal periodic regime, then specify the visual context and request schedules used in the implementation (Section~\ref{sec:rtc}).

\subsection{Action Coverage and Response Windows}
\paragraph{Latency determines the required action coverage.}
To analyze the timing constraint, consider a periodic steady regime with constant latency $L$ from a request to availability of its valid chunk.
Assume requests use fresh observations, the queue never runs empty, and there are no resets or additional scheduling and actuation delays.
Let $P$ be the duration of the remaining queue at request time, $C$ the duration of a new chunk, and $T$ the interval between requests.
The existing queue must cover inference, so $P\ge L$.
Because there is at most one request in flight, $T\ge L$; consuming every appended action without accumulating a growing queue requires $C=T$.
Thus,
\begin{equation}
  P\ge L,\qquad C=T\ge L,\qquad
  H_{\mathrm{plan}}=P+C\ge 2L.
  \label{eq:rtc-coverage}
\end{equation}
A request at time $t$ produces actions for $[t+P,t+P+C)$, so $H_{\mathrm{plan}}$ is the time from its observation to the end of its chunk.
The model must choose actions that remain effective over this horizon; it need not predict every future image.
Under these assumptions, the conditions guarantee continuous action coverage.

\paragraph{New observations can change actions only after the committed prefix.}
For an event $e$, let $v_e$ be the first time it is distinguishable from visual observations, and let $d_e$ be the latest time a corrective action can begin and still avoid the relevant failure.
The deadline is defined for the current situation and available controls, accounting for the actions executed before intervention.
Its \emph{available response window} is
\begin{equation}
  W(e)=d_e-v_e.
  \label{eq:response-window}
\end{equation}
For example, if a collision is one second away and avoiding it takes 0.3 seconds, the corrective action must begin within 0.7 seconds.

Consider an event that could not be inferred from earlier observations, requires a different action from the committed plan, and remains visible to the next request.
Let $s_e$ be the first request time that includes evidence of the event.
Under the periodic assumptions above, the first action that can depend on this evidence starts at $s_e+P$.
Its delay from the event becoming visible satisfies
\begin{equation}
  D(e)=s_e+P-v_e,\qquad P\le D(e)<P+T.
  \label{eq:response-delay}
\end{equation}
The event may arrive just before a request or just after one, which accounts for the additional wait of up to $T$.
This distinction between inference latency and the inference-cycle interval also appears in analyses of action-chunking policies~\citep{lu2026faster}.
A timely correction based on the new evidence requires $W(e)\ge D(e)$.
In the tight configuration $P=C=T=L$, the first response opportunity lies in $[L,2L)$; a response window of at least $2L$ covers every arrival phase in this idealized model.
Predictable changes can be anticipated, and already queued actions may remain appropriate without a correction.

\paragraph{Configured delay and measured latency are distinct.}
Our experiments use a response floor $F$: a reply completed in $\ell_i$ seconds is released after approximately $\max(\ell_i,F)$ seconds, before any further wait for action dispatch.
The floor is a lower bound, so it cannot replace the fixed latency $L$ in the analysis.
Actual queue coverage also depends on frame age, tick alignment, latency variation, and reply validity.
Longer delays require longer action coverage and postpone the point at which a new observation can affect execution; RTC makes this tradeoff explicit.

\subsection{Visual Context and Request Scheduling}
The visual context combines dense recent frames with sparsely sampled older frames.
Our implementation selects up to seven frames at tick offsets $(-90,-45,-15,-5,-2,-1,0)$, omitting unavailable history after startup or reset.
At 200\,ms per tick, these samples span up to 18 seconds, with the newest four covering the most recent second.
Each selected image carries observation timing; repeated samples may refer to the same captured image if no newer frame has arrived.
The direct-controller context also contains the most recent 120 input or reset events, compacting consecutive identical inputs, and the committed queue with action sequence identifiers.
Current tick, request time, episode deadline, and the previous protocol error complete the timing and execution context.
These fields describe the agent's own interaction, rather than privileged state of the game.

Each call is a newly constructed system--user request, with the task and action contract in the system message and visual and execution evidence in the user message.
Previous model reasoning is not carried forward.
The input records describe what has actually been dispatched or failed to dispatch; the queue describes actions still awaiting execution.
This distinction prevents an intended action from being treated as an observed game interaction.
In the direct-controller experiments, a new request is issued when the queue reaches the configured refill threshold and no request is in flight or reset is pending.

The initial trajectory collector uses the same visual sampling and append-only queue, but checks for a new request every 15 ticks, with no artificial response floor.
It requests 15-tick chunks for at most 60 seconds and 20 requests, subject to actual service latency.
The three-second sampling interval in collection is thus distinct from the three-second response floor used in direct-controller evaluation.
The resulting public bundle contains visual observations and executed-input and timing records, excluding raw model replies and private evaluator scores.
It is collected once per task and shared unchanged across development sessions and collaborative branches.

\FloatBarrier
\section{Peer Evidence and Changes in Development Direction}
\label{app:development-cases}

The collaborative evaluation records two distinct uses of peer evidence: adopting a working alternative and correcting an existing control rule.
For each case, we trace the prior approach, the evidence inspected or received, the submitted code change, and its execution outcome.
The cases are drawn from existing runs and selected because their use of peer evidence is traceable.
Developers receive public screenshots, input and timing records, and runtime feedback; private evaluator scores are used only in this analysis.

\subsection{Snake: Learning by Reviewing a Successful Peer}

The task requires collecting at least six apples before crashing.
In one independent development session, two submissions reach two and three apples before the output-token budget is exhausted.
The intervening and subsequent offline work repeatedly revisits head--tail identification, turn timing, and pixel thresholds; the final edits receive no third online test.
This session illustrates repeated refinement of a local control hypothesis with few further gameplay checks.

Within the collaborative team, branch D also struggles with a pixel-velocity controller.
Its fourth version adds direction voting and safety margins but reaches only one apple.
During review, D reads A's submitted policy, execution feedback, and screenshots.
It identifies A's grid occupancy tracking and search as more suitable than its own short-range extrapolation, and explicitly plans to adopt them.
The next submission implements a $17\times15$ grid estimated from public images, head tracking from changes in occupied cells, breadth-first search to the apple, and flood-fill evaluation of fallback moves.
The code changes therefore affect both state estimation and action selection.

\begin{table}[ht]
  \centering
  \small\setlength{\tabcolsep}{9pt}\renewcommand{\arraystretch}{1.12}
  \begin{tabular}{lrrrr|rrrr}
    \toprule
    Submission & 1 & 2 & 3 & 4 & 5 & 6 & 7 & 8 \\
    \midrule
    A: apples & 1 & 2 & 10 & 6 & --- & --- & --- & --- \\
    \rowcolor{metricband}
    D: apples & 0 & 0 & 2 & 1 & 6 & 6 & 4 & 6 \\
    \bottomrule
  \end{tabular}
  \caption{\textbf{Snake outcomes around the exchange.} Each value is the largest apple count reached before a crash within that submission's 60-second execution, which may contain resets. Exchange occurs between submissions 4 and 5. D inspects A's fourth submitted policy. Dashes mark branch-A submissions outside this case record.}
  \label{tab:snake-case}
\end{table}

\begin{figure}[ht]
  \centering
  \includegraphics[width=\linewidth]{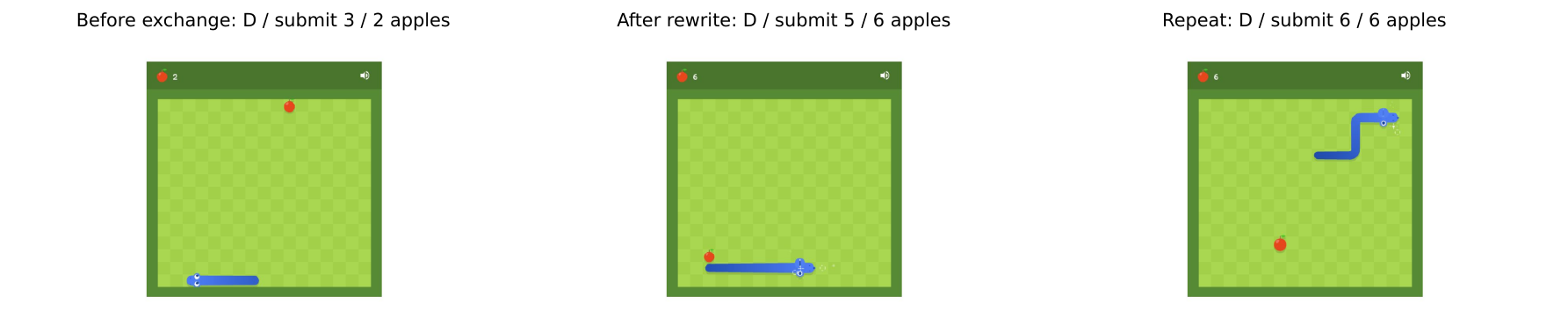}
  \caption{\textbf{Recorded visual evidence for Snake branch D.} The HUD shows two apples in submission 3, six in submission 5, and six again in submission 6. Frames are selected from each execution.}
  \label{fig:snake-case-rgb}
\end{figure}

The transfer occurs through D's own inspection of A, since D receives no valid incoming review from C.
A has already reached the goal on submission 3, so the exchange transfers a successful approach to a stalled branch.
D subsequently tries an additional lookahead modification, regresses to four apples, and restores the previous approach for its final successful submission.
Both the transfer and the regression are visible in Figure~\ref{fig:snake-development}.

\FloatBarrier
\subsection{NS-Shaft: Correcting a Persistent Waiting Rule}

The task requires surviving for at least 15 seconds without game over.
Branch A's fourth policy retains a rule that waits when a lower target platform aligns horizontally with the player, expecting a direct fall.
D's review identifies a conflict between that rule and the player still being supported by the current platform, and recommends moving to an edge before attempting the descent.
On resuming, A first reads the incoming review and explicitly acknowledges that waiting while supported is a weakness of its approach.
Its fifth submitted program removes the two aligned-target waiting branches and adds a standing-state rule that selects a direction to leave the support; the change appears in the executed policy, not only in the developer's stated plan.

\begin{figure}[ht]
  \centering
  \includegraphics[width=.85\linewidth]{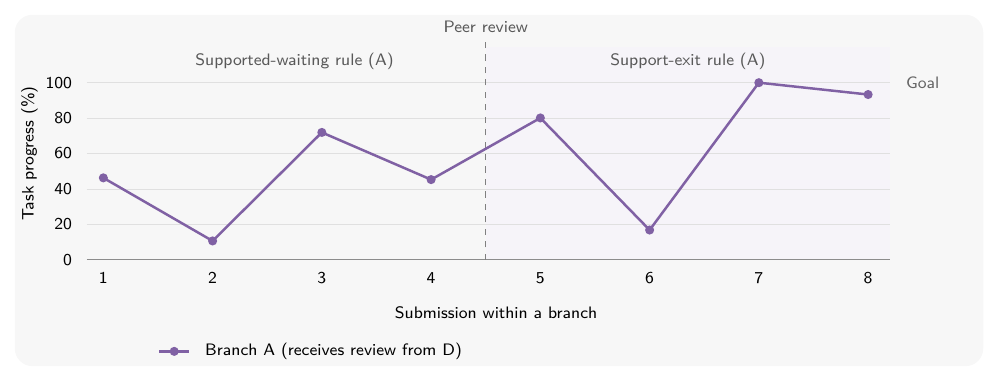}
  \caption{\textbf{A critique changes the NS-Shaft control rule.} A replaces the supported-player waiting rule after reviewing D's critique. The next submission improves progress, but later revisions include a regression before success at submission 7. Values are individual submission scores, equal to survival time divided by the 15-second goal and capped at 100\%; the vertical boundary marks peer review.}
  \label{fig:nsshaft-development}
\end{figure}

The longest survival increases from 6.793 seconds on submission 4 to 12.014 seconds on submission 5.
Further perception and control changes regress to 2.525 seconds on submission 6.
After correcting oscillatory steering around hazards, submission 7 reaches 18.330 seconds; a final adjustment yields 13.997 seconds on submission 8.
The peer critique thus precedes a concrete correction, while task completion also depends on later debugging.

\begin{figure}[ht]
  \centering
  \includegraphics[width=.85\linewidth]{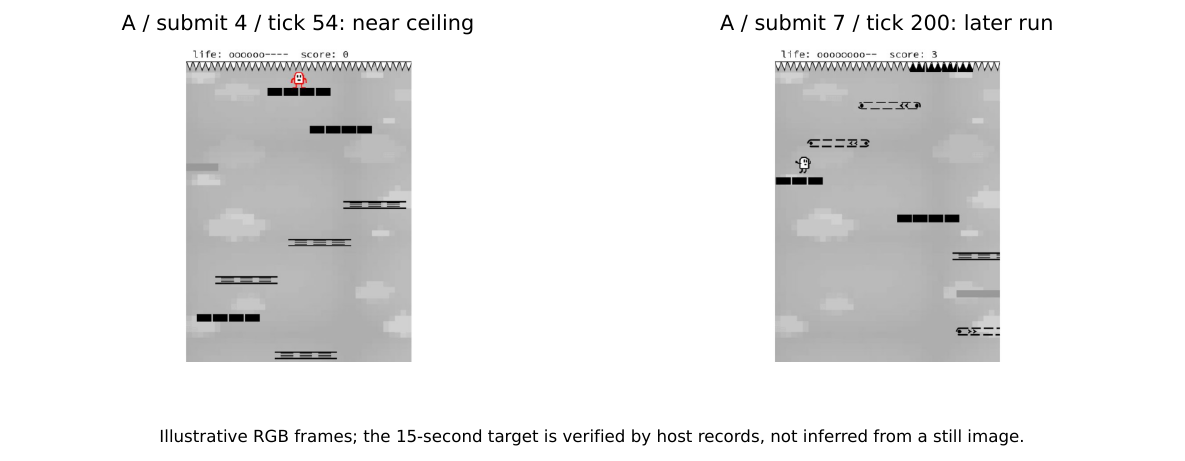}
  \caption{\textbf{NS-Shaft scenes before and after revision.} Selected frames show the supported player near the top of the scene in submission 4 and later gameplay in submission 7. Survival durations come from the evaluator's recorded run summaries.}
  \label{fig:nsshaft-case-rgb}
\end{figure}

\end{document}